\documentclass{article}

     \usepackage[final]{tackling_climate_workshop_style}
\usepackage{natbib}

\usepackage[utf8]{inputenc} 
\usepackage[T1]{fontenc}    
\usepackage{hyperref}       
\usepackage{url}            
\usepackage{booktabs}       
\usepackage{amsfonts}       
\usepackage{nicefrac}       
\usepackage{microtype}      
\let\tmlrAND\AND
\let\AND\relax
\usepackage{algorithm}
\usepackage{algorithmic}

\let\AND\tmlrAND
\usepackage{etoolbox}
\AtBeginEnvironment{algorithmic}{\let\AND\algoAND}
\usepackage{amsmath}
\usepackage{amssymb}
\usepackage{graphicx}
\usepackage{subcaption}
\usepackage{multirow}
\usepackage{bm}
\usepackage{algorithm}
\usepackage{algorithmic}
\usepackage{booktabs}
\usepackage{longtable}
\usepackage{amsthm}
\usepackage{mathtools}
\usepackage{enumitem}
\usepackage{graphicx}
\usepackage{tikz}

\title{Operator-Aware Posterior Sampling for Multivariate Weather Downscaling}

\author{%
  Carlo Saccardi \qquad Rudolf Saathof \qquad Geethu Joseph \qquad Justin Dauwels \\
  Delft University of Technology, The Netherlands \\
  \texttt{\{C.Saccardi, R.Saathof, G.Joseph, J.H.G.Dauwels\}@tudelft.nl}
}

\begin{document}

\maketitle

\begin{abstract} Statistical downscaling is the task of reconstructing high-resolution meteorological fields from coarse inputs. A principled approach formulates downscaling as a Bayesian inverse problem, using a diffusion prior over high-resolution fields and incorporating the coarse field only at inference through a likelihood. This decouples the prior from the coarse source, allowing one pretrained model to adapt to different sources. However, two challenges limit this flexibility: (1) when coarse and target fields come from different reanalysis systems, no fixed coarsening operator can capture their relationship; and (2) guidance gradients can differ by orders of magnitude across variables, making a single guidance strength ineffective.
This paper addresses both challenges within the guidance loop, leaving the prior unchanged. We learn the forward operator from paired data and normalize the gradient per channel. We show that this normalization corresponds to steepest descent under a block-max norm, allowing a single guidance strength across variables. Our code is publicly available at \url{https://github.com/CarloSaccardi/operator-aware-weather-downscaling}. \end{abstract}

\vspace{-2mm}
\section{Introduction}
\label{sec:intro}
\vspace{-1mm}
Kilometer-scale meteorological fields are essential for energy system planning \citep{QIU2024100263}, laser-satellite communications \citep{frehlich2010estimates}, and agriculture \citep{NeedforHighRes}. Global reanalyses such as ERA5 \citep{era5}, however, have a resolution of only $\sim$$25$\,km, leaving local terrain, coastlines, and other fine-scale features unresolved. Machine-learning weather models trained on these data inherit the same limitation \citep{Bodnar2025,Price2025,Bi2023}. Dynamical downscaling recovers these fine scales by running a regional model on a nested grid \citep{salathe2010regional}, at a computational cost that scales rapidly with the domain size \citep{Bader2009ClimateModels}. Statistical downscaling instead learns the coarse-to-fine mapping from paired data \citep{Wilby1998}, increasingly using diffusion models \citep{Mardani2025,merizzi2024wind}.

\begin{figure}[t]
    \centering
    \begin{subfigure}[b]{0.24\linewidth}
        \centering
        \includegraphics[width=\linewidth]{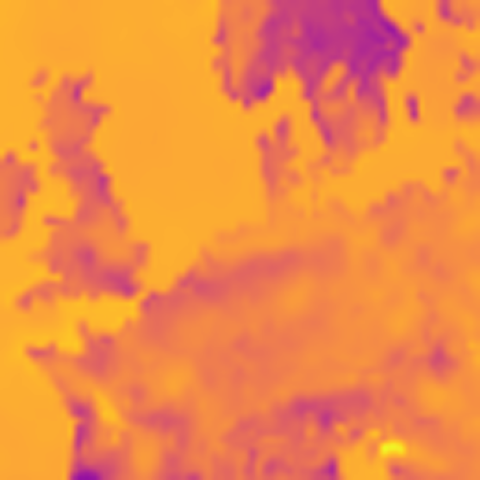}
        \caption{ERA5 input - sshf}
        \label{fig:era5_sshf}
    \end{subfigure}
    \hfill
    \begin{subfigure}[b]{0.24\linewidth}
        \centering
        \includegraphics[width=\linewidth]{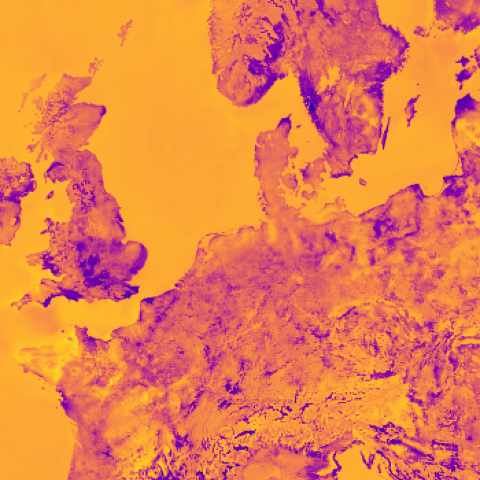}
        \caption{CERRA target - sshf}
        \label{fig:cerra_sshf}
    \end{subfigure}
    \hfill
    \begin{subfigure}[b]{0.24\linewidth}
        \centering
        \includegraphics[width=\linewidth]{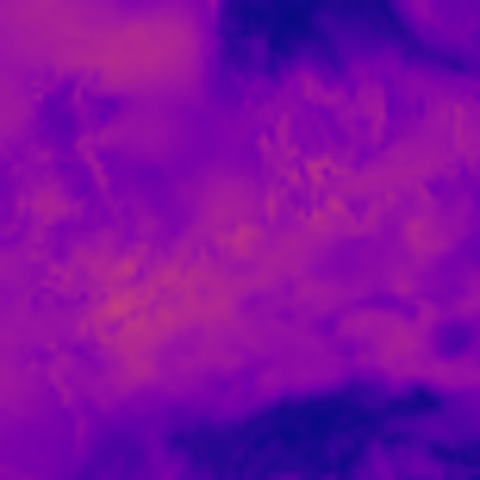}
        \caption{ERA5 input - zust}
        \label{fig:era5_zust}
    \end{subfigure}
    \hfill
    \begin{subfigure}[b]{0.24\linewidth}
        \centering
        \includegraphics[width=\linewidth]{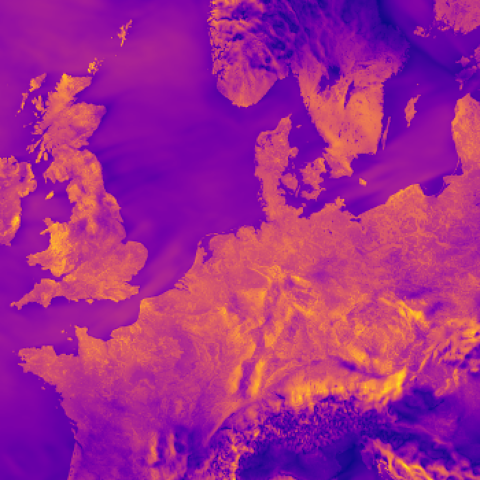}
        \caption{CERRA target - zust}
        \label{fig:cerra_zust}
    \end{subfigure}
    
    \caption{Comparison of \textit{surface sensible heat flux} (sshf) and \textit{friction velocity} (zust) between the coarse ERA5 and the high-resolution CERRA. Beyond the resolution gap, the fields differ in structure and large-scale pattern, implying an unknown and system-dependent coarse-to-fine map.}
    \label{fig:mismatch} 
\end{figure}

A complementary line of work treats downscaling as an inverse problem: a diffusion model trained on high-resolution fields as a prior and incorporating the coarse field at inference through a likelihood term, without retraining the prior \citep{chung2022diffusion,rozet2023score,tie2026zero,aich2026wind,schmidt2025generative}. This separates the expensive component, the prior over high-resolution atmospheric states, allowing a single pretrained prior to be adapted to different sources without retraining. Existing inverse problem formulations, however, assume a forward model $\mathbf{y} = \mathbf{A}(\mathbf{x}_0) + \mathbf{n}$ with $\mathbf{A}$ known, obtained by synthetically coarsening the high-resolution data \citep{harder2023hard,yang2024fourier}. This assumption fails when the coarse and fine fields come from different systems. For example, two reanalyses, or a climate model and a reanalysis, introduce systematic, terrain-dependent differences that cannot be captured by a coarsening operator (Figure~\ref{fig:mismatch}).

We address this limitation, together with a second challenge arising in multivariate guidance, entirely within the guidance loop of diffusion posterior sampling (DPS) \citep{chung2022diffusion}. First, we replace the assumed operator with a lightweight network $\mathbf{A}_\phi$ that learns the coarse-to-fine relationship from data. Mapping high-resolution to low-resolution is far easier than the reverse, so this network is cheap to train, and conditioning it on orography lets it capture the terrain-dependent bias between the two systems. Second, jointly downscaled variables have guidance gradients that differ by orders of magnitude, so one guidance strength overcorrects the rough variables and barely moves the smooth ones. We normalize the gradient per channel, which we show is steepest descent on the DPS objective under a block-max norm. Both modifications leave the prior unchanged. Adapting a new source requires only a two-layer network and one validated scalar \citep{aich2026wind, tie2026zero}, which we verify by refitting only $\mathbf{A}_\phi$ to a coarser source (App.~\ref{app:transfer}). On ERA5-to-CERRA, this trails conditional models trained end-to-end on a fixed coarse source, but does so without retraining the high-resolution prior. The framework therefore trades source-specific optimization for reuse of a single pretrained prior across heterogeneous coarse products.

\vspace{-2mm}
\section{Related Work}
\label{sec:relatedwork}
\vspace{-1mm}
Statistical downscaling learns a coarse-to-fine mapping from paired data, with deterministic \citep{harilal2022enhancedsd,Hohlein2020WindDownscaling, Saccardi2025} and, more recently, diffusion-based \citep{fotiadis2025adaptive,brenowitz2025climate,Mardani2025} models; much of this work synthetically coarsens the target to form its inputs, making the operator a known degradation \citep{harder2023hard,yang2024fourier}. A complementary line treats downscaling as an inverse problem, using a diffusion prior and incorporating the coarse field at inference through a likelihood \citep{tie2026zero,aich2026wind,schmidt2025generative}. This requires a forward operator, assumed known and typically a fixed downsampler, valid only when the coarse field is a degraded target. When the two come from different systems no such operator exists. Prior work debiases the coarse field and then trains a conditional model on the corrected pairs \citep{wan2023debias}, which handles the same problem but ties the model to the coarse source it was trained on. We instead keep the inverse-problem setup and learn the operator inside the guidance loop. Finally, DPS scales the guidance gradient by the residual norm with one scalar per sample \citep{chung2022diffusion}, which sets the step size but treats every variable alike; our per-channel normalization rescales each variable independently.

\vspace{-2mm}
\section{Background}
\label{sec:background}
\vspace{-1mm}
An inverse problem recovers a high-resolution signal $\mathbf{x}_0$ from a measurement $\mathbf{y}$ related to it by a forward operator $\mathbf{A}$, as $\mathbf{y} = \mathbf{A}(\mathbf{x}_0) + \mathbf{n}$ with noise $\mathbf{n} \sim \mathcal{N}(\mathbf{0}, \sigma^2 \mathbf{I})$. When $\mathbf{y}$ is lower-dimensional than $\mathbf{x}_0$ the problem is ill-posed, so rather than a point estimate a Bayesian formulation targets the posterior $p(\mathbf{x}_0 \mid \mathbf{y}) \propto p(\mathbf{y} \mid \mathbf{x}_0)\, p(\mathbf{x}_0)$, where the likelihood $p(\mathbf{y} \mid \mathbf{x}_0)$ is fixed by the forward model and a diffusion model \citep{ho2020denoising,song2020score} supplies the prior $p(\mathbf{x}_0)$.

To sample the posterior, DPS \citep{chung2022diffusion} adds a guidance correction to each step of the diffusion sampler, given by the gradient of the likelihood evaluated at the denoised estimate $\hat{\mathbf{x}}_0$ of $\mathbf{x}_0$,
\begin{equation}
\zeta_i \, \nabla_{\mathbf{x}_t} \big\| \mathbf{y} - \mathbf{A}(\hat{\mathbf{x}}_0) \big\|_2^2,
\qquad
\zeta_i = \zeta / \big\| \mathbf{y} - \mathbf{A}(\hat{\mathbf{x}}_0) \big\|_2 ,
\label{eq:dps_step}
\end{equation}
which steers the current estimate toward consistency with $\mathbf{y}$. The operator $\mathbf{A}$ maps a high-resolution field to its coarse counterpart and thus determines how the coarse observation influences the sampling process. The scalar $\zeta_i$ normalizes the update by the residual norm, controlling its overall magnitude while applying one common scaling factor to all components of the gradient.

\vspace{-2mm}
\section{Method}
\label{sec:method}
\vspace{-1mm}

In this section, we introduce our two contributions to the guidance loop: a learned forward operator that replaces the assumed downsampler with one fitted from data (Section~\ref{sec:operator}), and a per-channel normalization of the guidance gradient that lets a single guidance strength serve all variables at once (Section~\ref{sec:normalization}). Algorithm~\ref{alg:dps_chan} summarizes the full sampling procedure.

\vspace{-1mm}
\subsection{Learning the forward operator}
\label{sec:operator}
\vspace{-1mm}
The guidance correction in \eqref{eq:dps_step} requires a forward operator, and the standard choice is a fixed linear downsampler $\mathbf{A}$. Writing the residual it produces at the denoised estimate $\hat{\mathbf{x}}_0$ reveals why this is inadequate between different systems,
\begin{equation}
\mathbf{y} - \mathbf{A}\hat{\mathbf{x}}_0 \;=\; \underbrace{\mathbf{y} - \mathbf{A}\mathbf{x}_0}_{\text{inter-system bias } \bm{\delta}} \;+\; \underbrace{\mathbf{A}(\mathbf{x}_0 - \hat{\mathbf{x}}_0)}_{\text{reconstruction error}}.
\label{eq:residual_decomp}
\end{equation}
The reconstruction error vanishes as $\hat{\mathbf{x}}_0 \!\to\! \mathbf{x}_0$, but the bias $\bm{\delta} = \mathbf{y} - \mathbf{A}\mathbf{x}_0$ is independent of $\hat{\mathbf{x}}_0$ and is nonzero whenever $\mathbf{A}\mathbf{x}_0 \neq \mathbf{y}$, precisely the case when $\mathbf{y}$ and $\mathbf{x}_0$ come from different systems. The guidance gradient thus does not vanish at the target, steering the sample away from $\mathbf{x}_0$ toward a field whose coarsening matches $\mathbf{y}$. We remove this bias by replacing $\mathbf{A}$ with a small network $\mathbf{A}_\phi$ fitted directly to reproduce the coarse field,
\begin{equation}
\phi^\star = \arg\min_\phi \; \mathbb{E}_{\mathbf{x}_0}\big\| \mathbf{y} - \mathbf{A}_\phi(\mathbf{x}_0) \big\|_2^2 ,
\label{eq:operator_fit}
\end{equation}
which minimizes exactly the bias $\bm{\delta}$ a fixed operator leaves behind. At inference $\phi$ is frozen.

\vspace{-1mm}
\subsection{Per-channel gradient normalization}
\label{sec:normalization}
\vspace{-1mm}
The second component concerns how the guidance is scaled across variables. DPS was designed for natural images, whose channels are statistically alike, so one scale $\zeta$ suits all of them. At noise level $\sigma$ it forms a single gradient from the residual pooled over all $C$ channels and adds it to the drift,
\begin{equation}
\mathbf{g} \;=\; \nabla_{\mathbf{x}_\sigma} \big\| \mathbf{y} - \mathbf{A}_\phi(\hat{\mathbf{x}}_0) \big\|_2 ,
\qquad
\mathbf{d} \;=\; \frac{\mathbf{x}_\sigma - \hat{\mathbf{x}}_0}{\sigma} \;+\; \zeta\, \mathbf{g},
\qquad
\hat{\mathbf{x}}_0 = D_\theta(\mathbf{x}_\sigma; \sigma).
\label{eq:dps_drift}
\end{equation}
Weather variables are not alike: a smooth field such as surface pressure and a rough, turbulent one such as friction velocity produce guidance gradients of very different magnitude. The per-channel gradient norm $\|\mathbf{g}_c\|_F$ scales with the roughness of variable $c$, through the denoiser Jacobian $\mathbf{J} = \partial \hat{\mathbf{x}}_0 / \partial \mathbf{x}_\sigma$ (App.~\ref{app:normalization}), and can differ by orders across the $C$ variables. This imbalance persists despite per-variable standardization, since it originates in the Jacobian rather than the data scale. Thus, a single $\zeta$ cannot fit all of them at once: it is too large for the chaotic variables, and too small for the smooth ones. The resulting update is then governed by whichever variables are loudest.

We remove this coupling by normalizing each channel to unit norm before scaling,
\begin{equation}
\hat{\mathbf{g}}_c \;=\; \frac{\mathbf{g}_c}{\|\mathbf{g}_c\|_F + \varepsilon},
\qquad
\mathbf{d} \;=\; \frac{\mathbf{x}_\sigma - \hat{\mathbf{x}}_0}{\sigma} \;+\; \zeta\, \hat{\mathbf{g}},
\label{eq:update}
\end{equation}
reducing over the spatial dimensions only, so every variable contributes a unit-norm step and $\zeta$ moves each by the same amount regardless of its raw gradient magnitude. This is not an ad-hoc rescaling: it is exactly steepest descent under the block-max norm $\|\mathbf{v}\|_{\infty,F} = \max_c \|\mathbf{v}_c\|_F$, whose geometry treats each channel independently, so the optimal step normalizes each channel on its own, $\mathbf{v}_c^\star = -\mathbf{g}_c/\|\mathbf{g}_c\|_F$ (App.~\ref{app:normalization}).

\vspace{-2mm}
\section{Experiments}
\label{sec:setup}
\vspace{-1mm}
We downscale ERA5 \citep{era5} ($\sim$$25$\,km) to CERRA \citep{cerra} ($\sim$$5.5$\,km) over two European domains, Central Europe and Iberia, using six surface variables: $10$\,m winds ($u_{10}$, $v_{10}$), $2$\,m temperature ($t_{2m}$), surface sensible heat flux (sshf), friction velocity (zust), and surface pressure (sp), with orography as a static channel. The prior $p_\theta(\mathbf{x}_0\mid\mathbf{h})$ and operator $\mathbf{A}_\phi$ \eqref{eq:operator_fit} are trained separately and frozen at inference. We isolate each contribution with one ablation: the \textbf{operator ablation} compares the learned $\mathbf{A}_\phi$ against the bicubic downsampler, and the \textbf{normalization ablation} compares our update \eqref{eq:update} against DPS \eqref{eq:dps_step}, tuning $\zeta$ separately per arm so neither reduces to a mistuned baseline. We report per-variable RMSE, CRPS, and power-spectral-density (PSD) plots. Full data, training, and baseline details are in Appendix~\ref{app:experiments}.

\vspace{-1mm}
\paragraph{Results.}
Tables~\ref{tab:ablation_operator} and~\ref{tab:ablation_norm} isolate the two contributions on both regions. The learned operator improves RMSE and CRPS on every variable over the bicubic operator, with the largest gains on the variables a fixed operator mishandles most (see Figure~\ref{fig:fixed_A_failure}): on Central Europe, zust RMSE falls from $0.199$ to $0.090$ and sp from $131.2$ to $92.05$, and Iberia shows the same pattern. For the normalization ablation, per-channel normalization keeps every variable controlled under a single $\zeta$, whereas the DPS scaling leaves the highest-magnitude variable, sp, badly under-served: its RMSE rises to $153.7$ on Central Europe and $161.6$ on Iberia, against $92.05$ and $106.7$ for ours. All reported numbers are point estimates whose $95\%$ bootstrap intervals are given in Appendix~\ref{app:baselines}. The improvements we highlight, the learned operator on the rough variables and per-channel normalization on sp, exceed those intervals and are statistically significant, whereas the near-ties are within them.

\vspace{-2mm}
\section{Conclusion}
\label{sec:conclusion}
\vspace{-1mm}
We revisited diffusion posterior sampling for weather downscaling and addressed two limitations of its guidance loop that arise when the coarse and fine fields come from different systems. One, a fixed downsampler cannot express the bias between systems, so we learned the operator from data. Two, a single guidance strength cannot serve variables whose gradients differ by orders of magnitude, so we normalized the gradient per channel, which we show is steepest descent under a block-max norm. On ERA5-to-CERRA each contribution improves accuracy, with the largest gains on the rough variables a fixed operator mishandles and on the high-magnitude field one guidance scale under-serves. Richer operator architectures and more variables are natural next steps.

\begin{table}[t]
\centering
\caption{\textbf{Operator ablation.} RMSE and CRPS per variable, both regions. Both rows use per-channel normalization and the same $\zeta$; only the likelihood operator differs. The learned operator improves every variable. Best per column in \textbf{bold}.}
\label{tab:ablation_operator}
\scriptsize
\setlength{\tabcolsep}{3pt}
\begin{tabular}{ll cc cc cc cc cc cc}
\toprule
\multirow{2}{*}{Region} & \multirow{2}{*}{Operator} & \multicolumn{2}{c}{$u_{10}$} & \multicolumn{2}{c}{$v_{10}$} & \multicolumn{2}{c}{$t_{2m}$} & \multicolumn{2}{c}{sshf} & \multicolumn{2}{c}{zust} & \multicolumn{2}{c}{sp} \\
\cmidrule(lr){3-4} \cmidrule(lr){5-6} \cmidrule(lr){7-8} \cmidrule(lr){9-10} \cmidrule(lr){11-12} \cmidrule(lr){13-14}
 & & RMSE & CRPS & RMSE & CRPS & RMSE & CRPS & RMSE & CRPS & RMSE & CRPS & RMSE & CRPS \\
\midrule
\multirow{2}{*}{C.\ Europe}
  & Bicubic $\mathbf{A}$      & 1.199 & 0.739 & 1.174 & 0.718 & 1.338 & 0.806 & 33.83 & 15.48 & 0.199 & 0.119 & 131.2 & 75.46 \\
  & Ours                      & \textbf{0.986} & \textbf{0.553} & \textbf{0.982} & \textbf{0.547} & \textbf{1.021} & \textbf{0.571} & \textbf{32.70} & \textbf{13.33} & \textbf{0.090} & \textbf{0.047} & \textbf{92.05} & \textbf{55.96} \\
\midrule
\multirow{2}{*}{Iberia}
  & Bicubic $\mathbf{A}$      & 1.362 & 0.827 & 1.372 & 0.828 & 1.809 & 1.152 & 41.18 & 20.43 & 0.174 & 0.095 & 156.1 & 88.86 \\
  & Ours                      & \textbf{1.197} & \textbf{0.678} & \textbf{1.217} & \textbf{0.684} & \textbf{1.377} & \textbf{0.824} & \textbf{40.23} & \textbf{18.28} & \textbf{0.096} & \textbf{0.052} & \textbf{106.7} & \textbf{63.28} \\
\bottomrule
\end{tabular}
\end{table}
 
\begin{table}[t]
\centering
\caption{\textbf{Normalization ablation.} RMSE and CRPS per variable, both regions. Both rows use the learned operator; only the guidance scaling differs, each arm at its own tuned $\zeta$. A single per-sample scale (DPS) leaves the high-magnitude variable sp badly under-served (up to $+67\%$ RMSE), whereas ours keeps every variable controlled. Best per column in \textbf{bold}.}
\label{tab:ablation_norm}
\scriptsize
\setlength{\tabcolsep}{3pt}
\begin{tabular}{ll cc cc cc cc cc cc}
\toprule
\multirow{2}{*}{Region} & \multirow{2}{*}{Guidance} & \multicolumn{2}{c}{$u_{10}$} & \multicolumn{2}{c}{$v_{10}$} & \multicolumn{2}{c}{$t_{2m}$} & \multicolumn{2}{c}{sshf} & \multicolumn{2}{c}{zust} & \multicolumn{2}{c}{sp} \\
\cmidrule(lr){3-4} \cmidrule(lr){5-6} \cmidrule(lr){7-8} \cmidrule(lr){9-10} \cmidrule(lr){11-12} \cmidrule(lr){13-14}
 & & RMSE & CRPS & RMSE & CRPS & RMSE & CRPS & RMSE & CRPS & RMSE & CRPS & RMSE & CRPS \\
\midrule
\multirow{2}{*}{C.\ Europe}
  & DPS \citep{chung2022diffusion} & \textbf{0.969} & \textbf{0.539} & \textbf{0.966} & \textbf{0.535} & 1.092 & 0.608 & \textbf{31.27} & \textbf{12.79} & 0.096 & 0.049 & 153.7 & 93.17 \\
  & Ours                           & 0.986 & 0.553 & 0.982 & 0.547 & \textbf{1.021} & \textbf{0.571} & 32.70 & 13.33 & \textbf{0.090} & \textbf{0.047} & \textbf{92.05} & \textbf{55.96} \\
\midrule
\multirow{2}{*}{Iberia}
  & DPS \citep{chung2022diffusion} & \textbf{1.172} & \textbf{0.661} & \textbf{1.186} & \textbf{0.662} & 1.413 & 0.842 & \textbf{38.22} & \textbf{17.28} & \textbf{0.096} & \textbf{0.051} & 161.6 & 91.88 \\
  & Ours                           & 1.197 & 0.678 & 1.217 & 0.684 & \textbf{1.377} & \textbf{0.824} & 40.23 & 18.28 & \textbf{0.096} & 0.052 & \textbf{106.7} & \textbf{63.28} \\
\bottomrule
\end{tabular}
\end{table}

\newpage

\section*{Acknowledgments}
This publication is part of the project DAILSCOM: Atmospheric Turbulence Informed Machine Learning for Laser Satellite Communications with file number 20617 of the research programme \textit{Open Technology Program 2023-2}, which is financed by the Dutch Research Council (NWO). This work is also part of the project Delft AI4WF: Delft Artificial Intelligence for Weather Forecast with file number 2024.023 of the research programme \textit{Computing time of national computer systems}, which is financed by NWO under the grant \url{https://doi.org/10.61686/VYGRS56933}.

\bibliographystyle{plainnat}
\bibliography{main}
\appendix

\section{DPS-EDM algorithm}

Algorithm~\ref{alg:dps_chan} gives the full sampling procedure: an EDM stochastic Heun solver with the learned operator in the likelihood and per-channel normalization of the guidance gradient (line~6). The EDM parameterization is convenient here because its denoiser $D_\theta$ predicts $\hat{\mathbf{x}}_0$ directly from the noisy state, so the guidance operator $\mathbf{A}_\phi$ can be applied to $\hat{\mathbf{x}}_0$ without a separate Tweedie conversion. The correction is applied at both the predictor and corrector stages.

\begin{algorithm}[H]
\caption{EDM sampling with a learned operator and per-channel guidance normalization}
\label{alg:dps_chan}
\begin{algorithmic}[1]
\REQUIRE $N,\ \mathbf{y},\ \mathbf{h},\ D_\theta,\ \mathbf{A}_\phi,\ \zeta,\ \rho$
\STATE $\sigma_i \gets \big( \sigma_{\max}^{1/\rho} + \tfrac{i}{N-1}(\sigma_{\min}^{1/\rho} - \sigma_{\max}^{1/\rho}) \big)^{\rho}, \quad \sigma_N \gets 0$
\STATE $\mathbf{x}_0 \sim \mathcal{N}(\mathbf{0}, \sigma_0^2 \mathbf{I})$
\FOR{$i = 0$ \textbf{to} $N-1$}
    \STATE $\hat{\mathbf{x}}_0 \gets D_\theta(\mathbf{x}_i; \sigma_i, \mathbf{h})$
    \STATE $\mathbf{g} \gets \nabla_{\mathbf{x}_i} \big\| \mathbf{y} - \mathbf{A}_\phi(\hat{\mathbf{x}}_0, \mathbf{h}) \big\|_2$
    \STATE $\hat{\mathbf{g}}_c \gets \mathbf{g}_c / (\|\mathbf{g}_c\|_F + \varepsilon)$ \quad for each channel $c$ \hfill $\triangleright$ \emph{per-channel, ours}
    \STATE $\mathbf{d}_i \gets \tfrac{\mathbf{x}_i - \hat{\mathbf{x}}_0}{\sigma_i} + \zeta\, \hat{\mathbf{g}}$
    \STATE $\mathbf{x}_{\mathrm{tent}} \gets \mathbf{x}_i + (\sigma_{i+1} - \sigma_i)\, \mathbf{d}_i$
    \IF{$i < N-1$}
        \STATE $\hat{\mathbf{x}}_0' \gets D_\theta(\mathbf{x}_{\mathrm{tent}}; \sigma_{i+1}, \mathbf{h})$
        \STATE $\mathbf{g}' \gets \nabla_{\mathbf{x}_{\mathrm{tent}}} \big\| \mathbf{y} - \mathbf{A}_\phi(\hat{\mathbf{x}}_0', \mathbf{h}) \big\|_2$, \quad $\hat{\mathbf{g}}'_c \gets \mathbf{g}'_c / (\|\mathbf{g}'_c\|_F + \varepsilon)$
        \STATE $\mathbf{d}_i' \gets \tfrac{\mathbf{x}_{\mathrm{tent}} - \hat{\mathbf{x}}_0'}{\sigma_{i+1}} + \zeta\, \hat{\mathbf{g}}'$
        \STATE $\mathbf{x}_{i+1} \gets \mathbf{x}_i + (\sigma_{i+1} - \sigma_i)\, \tfrac{\mathbf{d}_i + \mathbf{d}_i'}{2}$
    \ELSE
        \STATE $\mathbf{x}_{i+1} \gets \mathbf{x}_{\mathrm{tent}}$
    \ENDIF
\ENDFOR
\RETURN $\mathbf{x}_N$
\end{algorithmic}
\end{algorithm}

\section{Per-channel gradient normalization: derivation}
\label{app:normalization}

We give the full derivation summarized in Section~\ref{sec:normalization}. We first factor the per-channel gradient magnitude to expose why it is imbalanced across variables (App.~\ref{app:factor}), then state and prove that per-channel normalization is normalized steepest descent under the block-max norm, with DPS the corresponding statement under the Euclidean norm (App.~\ref{app:blockmax}).

\subsection{Factorization of the per-channel gradient}
\label{app:factor}

Let $\mathbf{x}_\sigma$ be the noisy state at level $\sigma$ and $\hat{\mathbf{x}}_0 = D_\theta(\mathbf{x}_\sigma; \sigma)$ the denoiser output. The DPS guidance gradient is $\mathbf{g} = \nabla_{\mathbf{x}_\sigma} \| \mathbf{y} - \mathbf{A}_\phi(\hat{\mathbf{x}}_0) \|_2$, and by the chain rule through $\hat{\mathbf{x}}_0$,
\begin{equation}
\mathbf{g} \;=\; -\,\mathbf{J}^\top \mathbf{A}_\phi'(\hat{\mathbf{x}}_0)^\top \mathbf{r},
\qquad
\mathbf{J} = \frac{\partial \hat{\mathbf{x}}_0}{\partial \mathbf{x}_\sigma},
\qquad
\mathbf{r} = \frac{\mathbf{y} - \mathbf{A}_\phi(\hat{\mathbf{x}}_0)}{\| \mathbf{y} - \mathbf{A}_\phi(\hat{\mathbf{x}}_0) \|_2},
\label{eq:app_chain}
\end{equation}
where $\mathbf{J}$ is the denoiser Jacobian and $\mathbf{r}$ the unit residual. The magnitude of $\mathbf{g}$ in a given channel is therefore governed by $\mathbf{J}$: the residual $\mathbf{r}$ is a unit vector, and the operator $\mathbf{A}_\phi'$ is a fixed low-pass map, so the term that varies sharply across variables is the denoiser Jacobian. For a smooth variable such as surface pressure, $\hat{\mathbf{x}}_0$ barely changes as $\mathbf{x}_\sigma$ is perturbed, so $\|\mathbf{J}_c\|$ is small; for a rough, high-frequency variable such as friction velocity, small perturbations of $\mathbf{x}_\sigma$ produce large changes in $\hat{\mathbf{x}}_0$, so $\|\mathbf{J}_c\|$ is large. The per-channel gradient magnitudes $\|\mathbf{g}_c\|_F$ inherit this spread and differ by orders across the $C$ variables. Since the DPS drift adds $\zeta \mathbf{g}$ with a single scalar $\zeta$, the displacement of each variable is proportional to $\|\mathbf{g}_c\|_F$: the roughest variables dominate the update while the smoothest are barely moved.

\subsection{Per-channel normalization is block-max steepest descent}
\label{app:blockmax}
Per-channel normalization is not an ad-hoc rescaling: it is the normalized steepest-descent direction on the DPS objective under the block-max norm. We contrast it below with the Euclidean-normalized step, and with DPS itself, which applies no normalization across channels at all. Normalized steepest descent on $L$ under a norm $\|\cdot\|$ takes the unit-ball direction that best aligns with the negative gradient,
\begin{equation}
\mathbf{v}^\star \;=\; \arg\min_{\|\mathbf{v}\| \le 1} \; \langle \mathbf{g}, \mathbf{v} \rangle ,
\label{eq:app_nsd}
\end{equation}
and different norms yield different steps. Partition $\mathbf{v}$ and $\mathbf{g}$ into the $C$ channel blocks, so that $\langle \mathbf{g}, \mathbf{v} \rangle = \sum_{c=1}^C \langle \mathbf{g}_c, \mathbf{v}_c \rangle$.
Under the block-max norm $\|\mathbf{v}\|_{\infty,F} := \max_c \|\mathbf{v}_c\|_F$, the unit ball is the product set $\{ \mathbf{v} : \|\mathbf{v}_c\|_F \le 1 \ \forall c \}$, so the constraint separates across channels. The objective is separable too, so \eqref{eq:app_nsd} splits into $C$ independent problems $\min_{\|\mathbf{v}_c\|_F \le 1} \langle \mathbf{g}_c, \mathbf{v}_c \rangle$. Each is solved by Cauchy--Schwarz, giving
\begin{equation}
\mathbf{v}_c^\star \;=\; -\,\frac{\mathbf{g}_c}{\|\mathbf{g}_c\|_F}, \qquad c = 1, \dots, C,
\label{eq:app_blockmax_sol}
\end{equation}
which is exactly the per-channel normalized direction of \eqref{eq:update}: every variable contributes a unit-norm step, independent of its raw gradient magnitude.
Under the Euclidean norm, by contrast, the unit ball $\|\mathbf{v}\|_2 \le 1$ does not separate across channels, and the steepest-descent direction is the joint normalization
\begin{equation}
\mathbf{v}^\star \;=\; -\,\frac{\mathbf{g}}{\|\mathbf{g}\|_2}, \qquad \|\mathbf{g}\|_2 = \Big( \sum_c \|\mathbf{g}_c\|_F^2 \Big)^{1/2}.
\label{eq:app_euclidean}
\end{equation}
Here every block is divided by the same scalar $1/\|\mathbf{g}\|_2$, so its step retains the raw per-channel magnitude $\|\mathbf{g}_c\|_F$ and the imbalance of the previous subsection survives. DPS is the unnormalized limit of the same picture. Substituting $\zeta_i = \zeta / \|\mathbf{y} - \mathbf{A}_\phi(\hat{\mathbf{x}}_0)\|_2$ into \eqref{eq:dps_step} and using $\nabla \|\mathbf{r}\|_2^2 = 2\|\mathbf{r}\|_2 \nabla \|\mathbf{r}\|_2$, the residual norm cancels and the update is a fixed step $2\zeta$ along $\mathbf{g}$ itself, as written in \eqref{eq:dps_drift}: the per-sample scalar sets the overall step size but performs no rescaling across channels. Whether the common factor is $1/\|\mathbf{g}\|_2$ or absent entirely, the ratio between channels is unchanged and the loudest variables dominate the update. Block-max normalization coincides with these steps only when all $\|\mathbf{g}_c\|_F$ are equal; whenever they differ, it gives each variable its own unit-norm step.

\paragraph{Consequence for $\zeta$.}
Under the Euclidean step \eqref{eq:app_euclidean} the displacement of variable $c$ is $\zeta \|\mathbf{g}_c\|_F / \|\mathbf{g}\|_2$, which depends on the channel and on $\sigma$ through the Jacobian imbalance of \eqref{eq:app_chain}, so no single $\zeta$ is simultaneously appropriate for all variables. Under the block-max step \eqref{eq:app_blockmax_sol} the displacement of every variable is exactly $\zeta$, independent of $c$ and of $\sigma$. The guidance strength therefore acts as one scale-free displacement budget shared across all channels, which is what allows a single validated $\zeta$ to serve the full multivariate field.

\section{Experimental details}
\label{app:experiments}

\paragraph{Data.}
CERRA \citep{cerra} is a pan-European regional reanalysis at $\sim$$5.5$\,km, to our knowledge the only kilometer-scale reanalysis with continental coverage. We pair it with ERA5 \citep{era5} over two domains, Central Europe and Iberia, on six surface variables: $10$\,m winds ($u_{10}$, $v_{10}$), $2$\,m temperature ($t_{2m}$), surface sensible heat flux (sshf), friction velocity (zust), and surface pressure (sp). ERA5 ($\sim$$25$\,km, $77\times77$) is regridded to the CERRA projection ($\sim$$5.5$\,km, $350\times350$), a $\sim$$4.55\times$ factor; the grid ratio $350/77$ is not an integer, so no exact area-average stencil exists. A CERRA-derived orography field is included as a static channel and all variables are standardized with per-variable ERA5 statistics. We train on 2013--2019 and the odd months of 2020, validate on the even months of 2020, and test on 2021. Unlike joint multi-region training, the prior and operator are trained \emph{separately per domain}: one model on Central Europe and one on Iberia, each evaluated on its own held-out test year.

\paragraph{Backbone and training.}
The prior is an EDM denoiser \citep{karras2022elucidating} on a SongUNet backbone (as implemented in NVIDIA PhysicsNeMo \citep{physicsnemo2023}) with base channel count $32$, channel multipliers $(1,2,2)$, and $4$ residual blocks per resolution. Its input concatenates the noised target ($6$ channels) and the orography ($1$ channel); it never sees ERA5. We train with Adam ($\beta_1{=}0.9$, $\beta_2{=}0.999$) at a constant learning rate $8\times10^{-4}$ in mixed precision. During training the noise level is drawn log-normally ($P_{\text{mean}}{=}0.0$, $P_{\text{std}}{=}1.2$, $\sigma_{\text{data}}{=}1.0$), with the network supported over $\sigma_{\min}{=}0.002$ to $\sigma_{\max}{=}800$. Sampling uses the stochastic Heun solver with $18$ steps, and all reported quantities use an $8$-member ensemble. $\mathbf{A}_\phi$ corrects an area-interpolation of $\mathbf{x}_0$ with a two-layer convolutional network ($3\times3$ kernels, $16$ hidden channels, GELU) that also sees the interpolated orography, fitted by \eqref{eq:operator_fit} on the same paired data and frozen at inference.

\paragraph{Guidance weight.}
The scalar $\zeta$ is selected by grid search on the validation split, minimizing the mean per-variable RMSE across the six standardized variables, and tuned separately for the per-channel and DPS arms because the two place $\zeta$ on different scales (a unit-norm direction versus a raw gradient). We sweep $\zeta \in \{\,0.01, 0.1, 0.5, 1, 10, 20, 50, 100, 200, 500, 1000\,\}$ for the per-channel arm and DPS, retaining in each case the value with the lowest validation error. The selected weights are $\zeta{=}10$ for the per-channel arm and $\zeta{=}200$ for DPS, identical on both regions.

\paragraph{Metrics.}
We report per-variable root mean square error (RMSE) and continuous ranked probability score (CRPS) of the $8$-member ensemble, both in physical units, as pointwise and probabilistic measures of accuracy. Neither penalizes a method for producing an over-smooth field that scores well on average yet misses the fine-scale structure downscaling is meant to recover, so we complement them with two distributional diagnostics. Isotropic power spectral density (PSD) plots (Figures~\ref{fig:psd_ce} and~\ref{fig:psd_iberia}) show how each method distributes energy across wavenumbers, revealing whether guidance restores the small-scale variability of the CERRA target or flattens it. Per-variable marginal distributions (Figures~\ref{fig:logpdf_ce} and~\ref{fig:logpdf_iberia}), plotted as $\log_{10}$ densities so the tails remain visible, show whether the full range of values, including rare extremes, is reproduced rather than concentrated near the mean.

\paragraph{Baselines.}
For context we also train three models that condition on the coarse field at training time rather than using it at inference: a conditional diffusion model (CondDiff), CorrDiff \citep{Mardani2025}, and a deterministic U-Net, all sharing the backbone above. Their results and per-variable marginal distributions are reported in Appendix~\ref{app:baselines}.

\section{Comparison with conditional baselines}
\label{app:baselines}

For context we compare against three models that condition on the coarse field at training time rather than using it at inference: a conditional diffusion model (CondDiff), CorrDiff \citep{Mardani2025}, and a deterministic U-Net, all sharing the backbone of Appendix~\ref{app:experiments}. Tables~\ref{tab:baselines_ce} and~\ref{tab:baselines_ib} report RMSE and CRPS on both regions. We quantify the sampling uncertainty of every entry by bootstrapping the test fields ($2000$ resamples in blocks of $24$); the resulting $95\%$ interval half-widths are given in Tables~\ref{tab:ci_ce} and~\ref{tab:ci_ib}, and cover the test draw only, not seed-to-seed variability. On sp, where the fixed-scale DPS baseline is worst, the gap to ours is far larger than either interval ($153.7 \pm 13.9$ vs.\ $92.05 \pm 3.6$ on Central Europe, $161.6 \pm 11.1$ vs.\ $106.7 \pm 5.2$ on Iberia), while on the remaining variables the methods overlap.

The conditional models are the more accurate here, which is expected: they are trained end-to-end to map the coarse field directly to the high-resolution target, so the coarse-to-fine relationship is learned by the full diffusion model rather than mediated by a small operator inside the likelihood. Our posterior-sampling approach remains competitive across variables, though less accurate than the conditional models. What it retains instead is the modularity of the inverse-problem framing: because the coarse field enters only through the learned operator $\mathbf{A}_\phi$, adapting the method to a different coarse source, for instance downscaling a climate model instead of ERA5, requires retraining only $\mathbf{A}_\phi$, a two-layer network, while the diffusion prior is reused unchanged (App.~\ref{app:transfer}). A conditional model, by contrast, is tied to the coarse source it was trained on and must be retrained in full when that source changes. Figures~\ref{fig:psd_ce} and~\ref{fig:psd_iberia} report the power spectral density of every model against the CERRA target, and Figures~\ref{fig:logpdf_ce} and~\ref{fig:logpdf_iberia} their marginal distributions, on both regions and all six variables.

%
%
%

\newcommand{\gpwidth}{0.32\linewidth}

\newlength{\gpht}

\newcommand{\rowlab}[1]{%
  \raisebox{0.5\gpht}{\rotatebox[origin=c]{90}{\small #1}}\,%
}

\newcommand{\gridpanel}[2]{%
  \begin{tabular}{@{}c@{}}
    \includegraphics[width=\gpwidth]{#1}\\[-2pt]
    {\small #2}
  \end{tabular}%
}

\newcommand{\gridgap}{\hspace{0.004\linewidth}}
\newcommand{\gridrowsep}{\vspace{3pt}}

\newcommand{\xlab}[1]{\vspace{1pt}\par{\small #1}}

\newcommand{\gridlegend}{%
  \par\vspace{3pt}{\centering\includegraphics{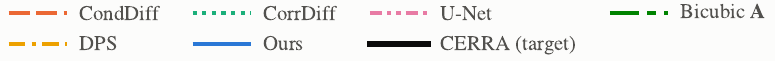}\par}\vspace{3pt}%
}

\setlength{\intextsep}{3pt}
\setlength{\abovecaptionskip}{4pt}


\begin{table}[H]
\centering
\caption{\textbf{Central Europe: full comparison.} Conditional baselines (top) condition on the coarse field at training time; the inference-time methods (bottom) use it through the likelihood, with Ours our full method. Best per column in \textbf{bold} (lower is better). $^{\dagger}$The U-Net is deterministic, so its CRPS is a single-prediction MAE, not comparable to the $8$-member ensemble CRPS; it is excluded from the CRPS comparison.}
\label{tab:baselines_ce}
\scriptsize
\setlength{\tabcolsep}{3pt}
\begin{tabular}{l cc cc cc cc cc cc}
\toprule
\multirow{2}{*}{Model} & \multicolumn{2}{c}{$u_{10}$} & \multicolumn{2}{c}{$v_{10}$} & \multicolumn{2}{c}{$t_{2m}$} & \multicolumn{2}{c}{sshf} & \multicolumn{2}{c}{zust} & \multicolumn{2}{c}{sp} \\
\cmidrule(lr){2-3} \cmidrule(lr){4-5} \cmidrule(lr){6-7} \cmidrule(lr){8-9} \cmidrule(lr){10-11} \cmidrule(lr){12-13}
 & RMSE & CRPS & RMSE & CRPS & RMSE & CRPS & RMSE & CRPS & RMSE & CRPS & RMSE & CRPS \\
\midrule
CondDiff                     & 0.904 & 0.489 & 0.903 & 0.486 & 0.843 & 0.452 & 27.08 & 10.22 & 0.075 & 0.039 & \textbf{59.98} & \textbf{35.56} \\
CorrDiff \citep{Mardani2025} & 0.884 & \textbf{0.477} & 0.877 & \textbf{0.471} & 0.831 & \textbf{0.449} & 25.76 & \textbf{9.782} & 0.072 & \textbf{0.038} & 72.01 & 41.82 \\
U-Net$^{\dagger}$            & \textbf{0.856} & 0.616 & \textbf{0.852} & 0.610 & \textbf{0.831} & 0.641 & \textbf{25.42} & 12.76 & \textbf{0.071} & 0.049 & 116.0 & 87.82 \\
\midrule
Bicubic $\mathbf{A}$         & 1.199 & 0.739 & 1.174 & 0.718 & 1.338 & 0.806 & 33.83 & 15.48 & 0.199 & 0.119 & 131.2 & 75.46 \\
DPS \citep{chung2022diffusion} & 0.969 & 0.539 & 0.966 & 0.535 & 1.092 & 0.608 & 31.27 & 12.79 & 0.096 & 0.049 & 153.7 & 93.17 \\
Ours                         & 0.986 & 0.553 & 0.982 & 0.547 & 1.021 & 0.571 & 32.70 & 13.33 & 0.090 & 0.047 & 92.05 & 55.96 \\
\bottomrule
\end{tabular}
\end{table}

\begin{table}[H]
\centering
\caption{\textbf{Central Europe: bootstrap intervals.} Each entry is the half-width ($\pm$) of the $95\%$ interval for the matching entry in Table~\ref{tab:baselines_ce}; the full interval is that table's value $\pm$ the half-width here. From $2000$ resamples of the test fields in blocks of $24$. The interval covers the test draw only; with a single training seed it says nothing about seed-to-seed variability. $^{\dagger}$U-Net CRPS is a single-member MAE (see Table~\ref{tab:baselines_ce}).}
\label{tab:ci_ce}
\scriptsize
\setlength{\tabcolsep}{3pt}
\begin{tabular}{l cc cc cc cc cc cc}
\toprule
\multirow{2}{*}{Model} & \multicolumn{2}{c}{$u_{10}$} & \multicolumn{2}{c}{$v_{10}$} & \multicolumn{2}{c}{$t_{2m}$} & \multicolumn{2}{c}{sshf} & \multicolumn{2}{c}{zust} & \multicolumn{2}{c}{sp} \\
\cmidrule(lr){2-3} \cmidrule(lr){4-5} \cmidrule(lr){6-7} \cmidrule(lr){8-9} \cmidrule(lr){10-11} \cmidrule(lr){12-13}
 & RMSE & CRPS & RMSE & CRPS & RMSE & CRPS & RMSE & CRPS & RMSE & CRPS & RMSE & CRPS \\
\midrule
CondDiff                     & 0.020 & 0.010 & 0.020 & 0.010 & 0.029 & 0.015 & 3.29 & 0.841 & 0.0018 & 0.0010 & 3.76 & 2.41 \\
CorrDiff \citep{Mardani2025} & 0.020 & 0.010 & 0.020 & 0.009 & 0.028 & 0.015 & 3.21 & 0.867 & 0.0016 & 0.0009 & 2.38 & 1.34 \\
U-Net$^{\dagger}$            & 0.019 & 0.012 & 0.019 & 0.012 & 0.026 & 0.018 & 3.11 & 1.06 & 0.0015 & 0.0011 & 3.09 & 2.40 \\
\midrule
Bicubic $\mathbf{A}$         & 0.035 & 0.023 & 0.031 & 0.019 & 0.045 & 0.031 & 3.32 & 1.06 & 0.0084 & 0.0054 & 9.20 & 5.15 \\
DPS \citep{chung2022diffusion} & 0.021 & 0.010 & 0.021 & 0.009 & 0.035 & 0.021 & 3.39 & 1.00 & 0.0036 & 0.0018 & 13.85 & 8.96 \\
Ours                         & 0.022 & 0.011 & 0.021 & 0.010 & 0.032 & 0.019 & 3.65 & 1.09 & 0.0023 & 0.0011 & 3.62 & 2.46 \\
\bottomrule
\end{tabular}
\end{table}

\begin{table}[H]
\centering
\caption{\textbf{Iberia: full comparison.} Layout and conventions as in Table~\ref{tab:baselines_ce}. $^{\dagger}$The U-Net CRPS is a single-member MAE and is excluded from the CRPS comparison.}
\label{tab:baselines_ib}
\scriptsize
\setlength{\tabcolsep}{3pt}
\begin{tabular}{l cc cc cc cc cc cc}
\toprule
\multirow{2}{*}{Model} & \multicolumn{2}{c}{$u_{10}$} & \multicolumn{2}{c}{$v_{10}$} & \multicolumn{2}{c}{$t_{2m}$} & \multicolumn{2}{c}{sshf} & \multicolumn{2}{c}{zust} & \multicolumn{2}{c}{sp} \\
\cmidrule(lr){2-3} \cmidrule(lr){4-5} \cmidrule(lr){6-7} \cmidrule(lr){8-9} \cmidrule(lr){10-11} \cmidrule(lr){12-13}
 & RMSE & CRPS & RMSE & CRPS & RMSE & CRPS & RMSE & CRPS & RMSE & CRPS & RMSE & CRPS \\
\midrule
CondDiff                     & 1.039 & 0.552 & 1.070 & 0.571 & 0.916 & 0.500 & 26.47 & 9.923 & 0.077 & 0.040 & \textbf{74.09} & \textbf{45.00} \\
CorrDiff \citep{Mardani2025} & 1.020 & \textbf{0.541} & 1.036 & \textbf{0.550} & 0.917 & \textbf{0.507} & 25.28 & \textbf{9.548} & 0.074 & \textbf{0.039} & 79.20 & 46.70 \\
U-Net$^{\dagger}$            & \textbf{0.998} & 0.702 & \textbf{1.015} & 0.714 & \textbf{0.916} & 0.733 & \textbf{25.08} & 12.30 & \textbf{0.073} & 0.050 & 127.4 & 98.46 \\
\midrule
Bicubic $\mathbf{A}$         & 1.362 & 0.827 & 1.372 & 0.828 & 1.809 & 1.152 & 41.18 & 20.43 & 0.174 & 0.095 & 156.1 & 88.86 \\
DPS \citep{chung2022diffusion} & 1.172 & 0.661 & 1.186 & 0.662 & 1.413 & 0.842 & 38.22 & 17.28 & 0.096 & 0.051 & 161.6 & 91.88 \\
Ours                         & 1.197 & 0.678 & 1.217 & 0.684 & 1.377 & 0.824 & 40.23 & 18.28 & 0.096 & 0.052 & 106.7 & 63.28 \\
\bottomrule
\end{tabular}
\end{table}

\begin{table}[H]
\centering
\caption{\textbf{Iberia: bootstrap intervals.} Each entry is the half-width ($\pm$) of the $95\%$ interval for the matching entry in Table~\ref{tab:baselines_ib}; the full interval is that table's value $\pm$ the half-width here. Resampling as in Table~\ref{tab:ci_ce}.}
\label{tab:ci_ib}
\scriptsize
\setlength{\tabcolsep}{3pt}
\begin{tabular}{l cc cc cc cc cc cc}
\toprule
\multirow{2}{*}{Model} & \multicolumn{2}{c}{$u_{10}$} & \multicolumn{2}{c}{$v_{10}$} & \multicolumn{2}{c}{$t_{2m}$} & \multicolumn{2}{c}{sshf} & \multicolumn{2}{c}{zust} & \multicolumn{2}{c}{sp} \\
\cmidrule(lr){2-3} \cmidrule(lr){4-5} \cmidrule(lr){6-7} \cmidrule(lr){8-9} \cmidrule(lr){10-11} \cmidrule(lr){12-13}
 & RMSE & CRPS & RMSE & CRPS & RMSE & CRPS & RMSE & CRPS & RMSE & CRPS & RMSE & CRPS \\
\midrule
CondDiff                     & 0.025 & 0.012 & 0.027 & 0.014 & 0.022 & 0.011 & 1.99 & 0.592 & 0.0017 & 0.0008 & 6.34 & 4.34 \\
CorrDiff \citep{Mardani2025} & 0.027 & 0.013 & 0.027 & 0.014 & 0.023 & 0.013 & 2.10 & 0.629 & 0.0017 & 0.0009 & 2.95 & 1.85 \\
U-Net$^{\dagger}$            & 0.025 & 0.016 & 0.027 & 0.017 & 0.023 & 0.016 & 2.00 & 0.767 & 0.0017 & 0.0011 & 4.96 & 4.08 \\
\midrule
Bicubic $\mathbf{A}$         & 0.031 & 0.020 & 0.028 & 0.017 & 0.039 & 0.029 & 2.54 & 1.11 & 0.0078 & 0.0034 & 10.99 & 5.70 \\
DPS \citep{chung2022diffusion} & 0.024 & 0.013 & 0.026 & 0.013 & 0.043 & 0.029 & 2.60 & 1.08 & 0.0024 & 0.0012 & 11.05 & 5.99 \\
Ours                         & 0.026 & 0.014 & 0.028 & 0.014 & 0.031 & 0.020 & 2.74 & 1.18 & 0.0023 & 0.0012 & 5.24 & 3.43 \\
\bottomrule
\end{tabular}
\end{table}

\clearpage

The spectra and marginals in Figures~\ref{fig:psd_ce} and~\ref{fig:logpdf_ce} tell a consistent story. Across the resolved scales every method overlies the CERRA target, so the differences appear only at the highest wavenumbers, and only for the rough variables. For \textit{sshf} and \textit{zust}, the fixed-operator and deterministic baselines (bicubic and U-Net) lose high-frequency energy and their marginals contract toward the mean, undershooting the target tails. Ours retains the most small-scale energy and reproduces these tails most closely, with DPS close behind. For the smooth variables ($u_{10}$, $v_{10}$, $t_{2m}$, sp) all methods track both the spectrum and the full marginal, with negligible separation. The large \textit{sp} error gap between our proposed per-channel gradient normalization  and DPS seen in the tables leaves no visible trace here, since it reflects a bias in the mean field rather than a mismatch in the spectrum or marginal distribution.

\gridlegend

\begin{figure}[H]
  \centering
  \rowlab{PSD}%
  \gridpanel{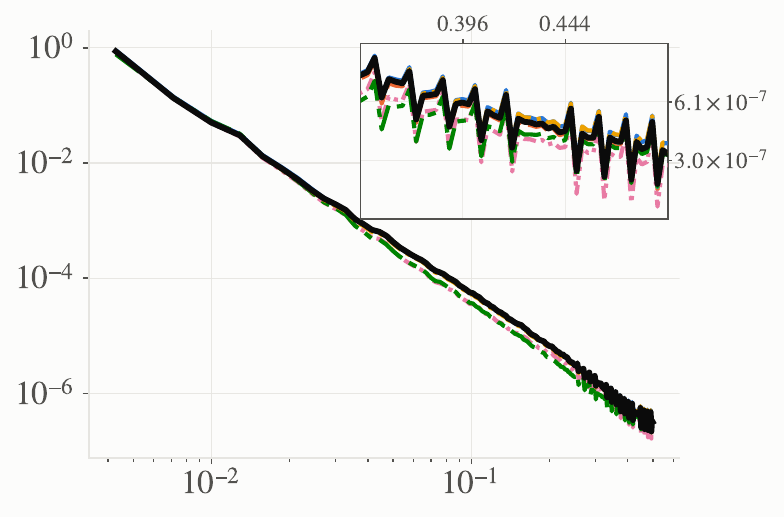}{$u_{10}$ $[\mathrm{m^{2}s^{-2}}]$}\gridgap
  \gridpanel{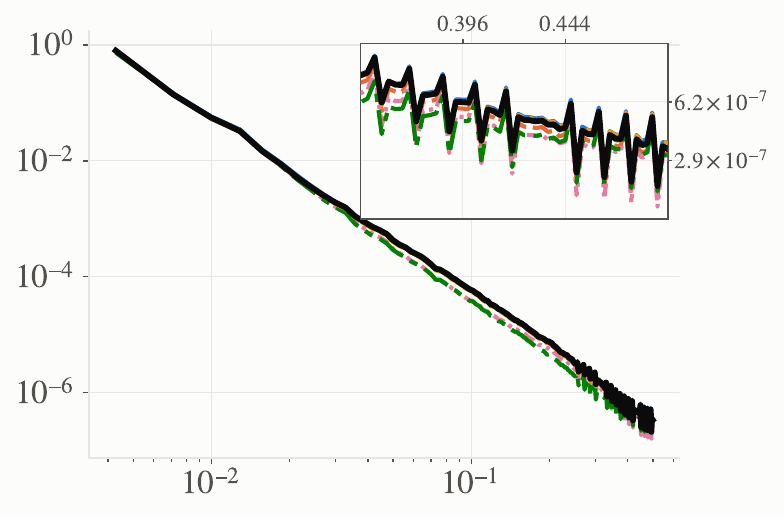}{$v_{10}$ $[\mathrm{m^{2}s^{-2}}]$}\gridgap
  \gridpanel{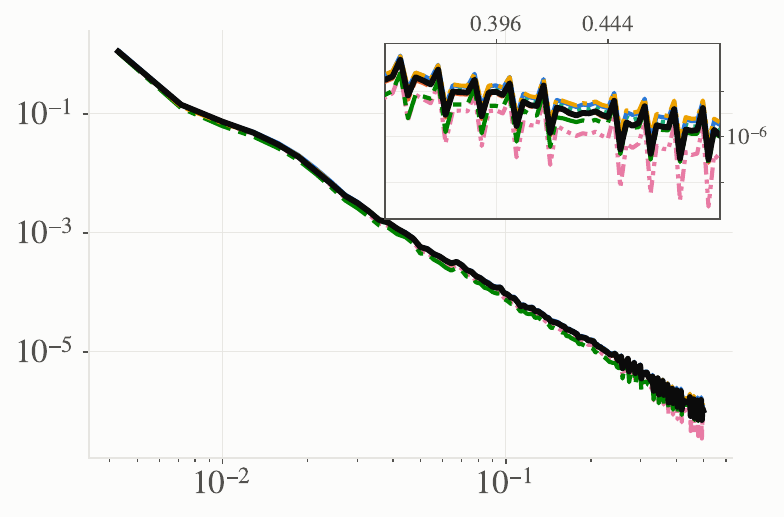}{$t_{2m}$ $[\mathrm{K^{2}}]$}

  \gridrowsep

  \rowlab{PSD}%
  \gridpanel{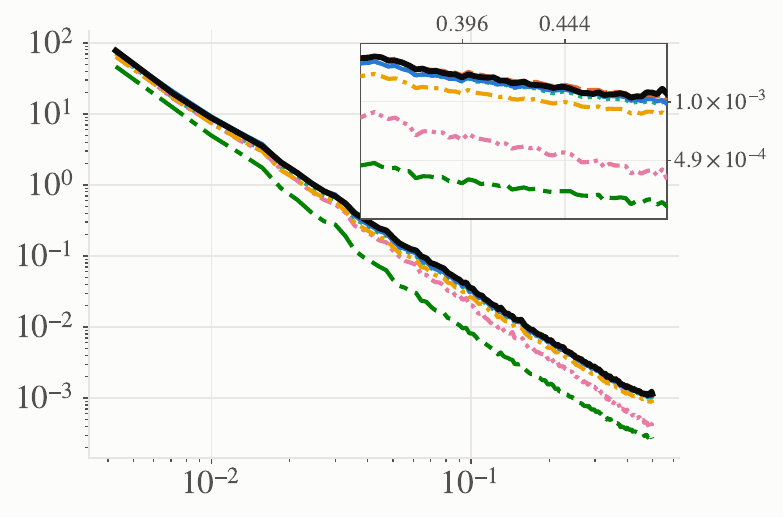}{sshf $[\mathrm{W^{2}m^{-4}}]$}\gridgap
  \gridpanel{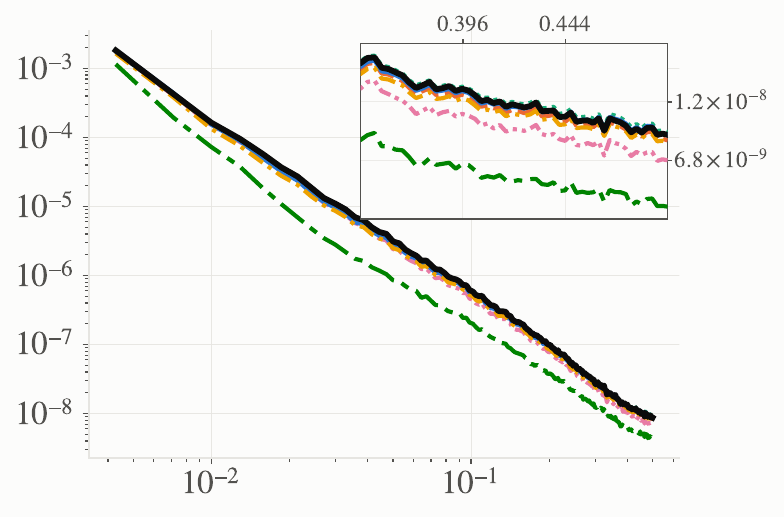}{zust $[\mathrm{m^{2}s^{-2}}]$}\gridgap
  \gridpanel{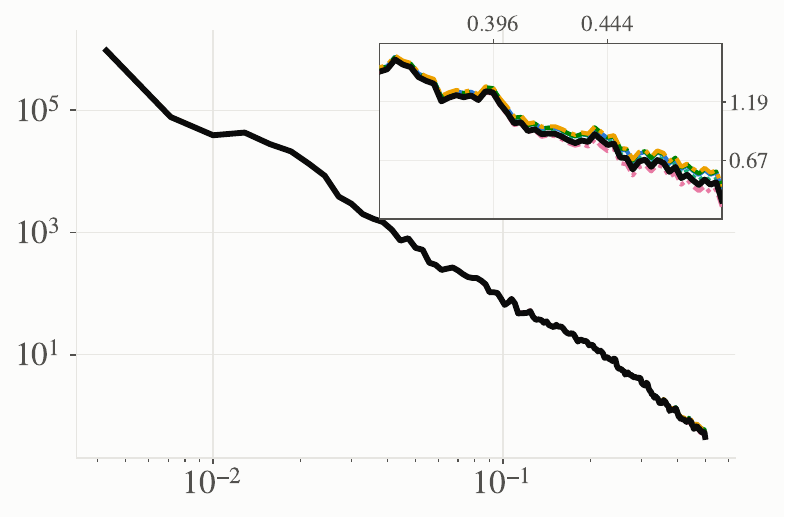}{sp $[\mathrm{Pa^{2}}]$}

  \xlab{wavenumber $k$ $[\mathrm{cycles\;cell^{-1}}]$}

  \caption{\textbf{Power spectral density, Central Europe.} Test split, 8-member ensembles. The inset magnifies the highest $0.15$ decades of
  wavenumber, where the curves separate.}
  \label{fig:psd_ce}
\end{figure}

\begin{figure}[H]
  \centering
  \rowlab{$\log_{10}$ PDF}%
  \gridpanel{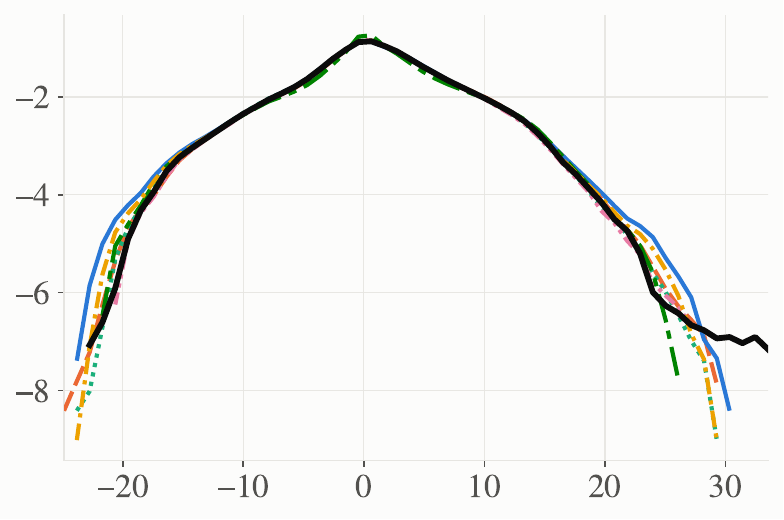}{$u_{10}$ $[\mathrm{m\,s^{-1}}]$}\gridgap
  \gridpanel{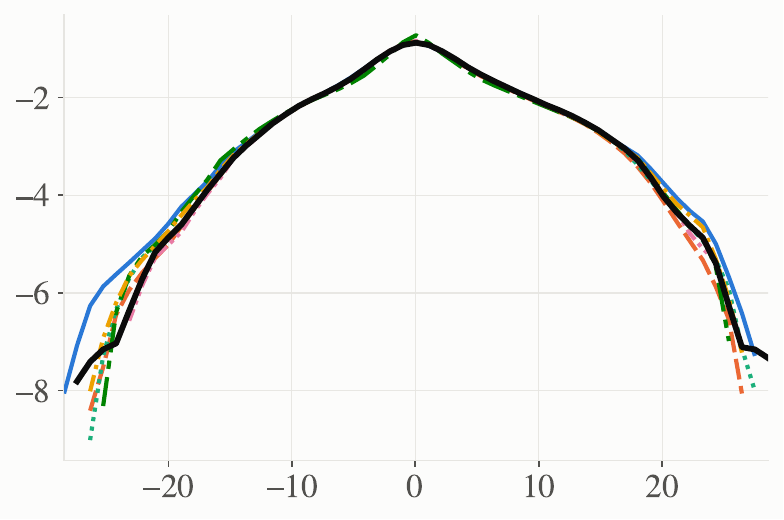}{$v_{10}$ $[\mathrm{m\,s^{-1}}]$}\gridgap
  \gridpanel{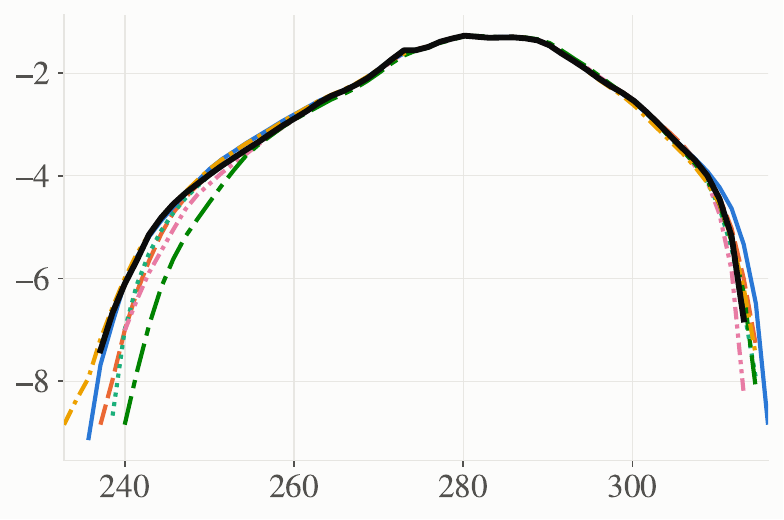}{$t_{2m}$ $[\mathrm{K}]$}

  \gridrowsep

  \rowlab{$\log_{10}$ PDF}%
  \gridpanel{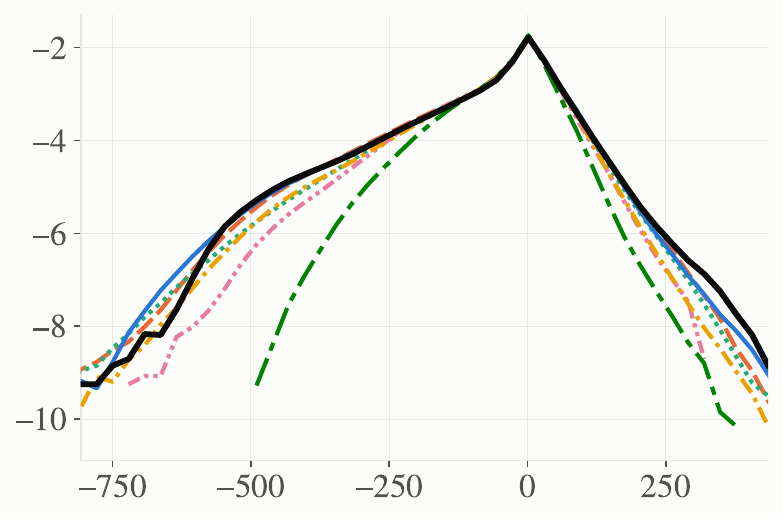}{sshf $[\mathrm{W\,m^{-2}}]$}\gridgap
  \gridpanel{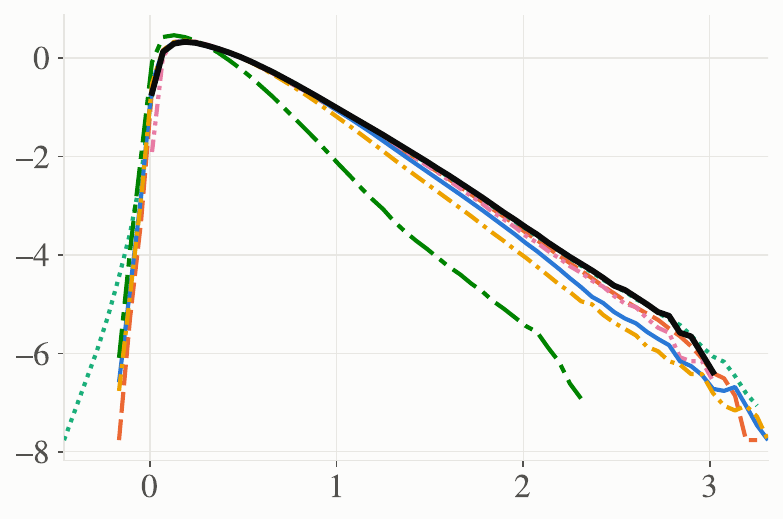}{zust $[\mathrm{m\,s^{-1}}]$}\gridgap
  \gridpanel{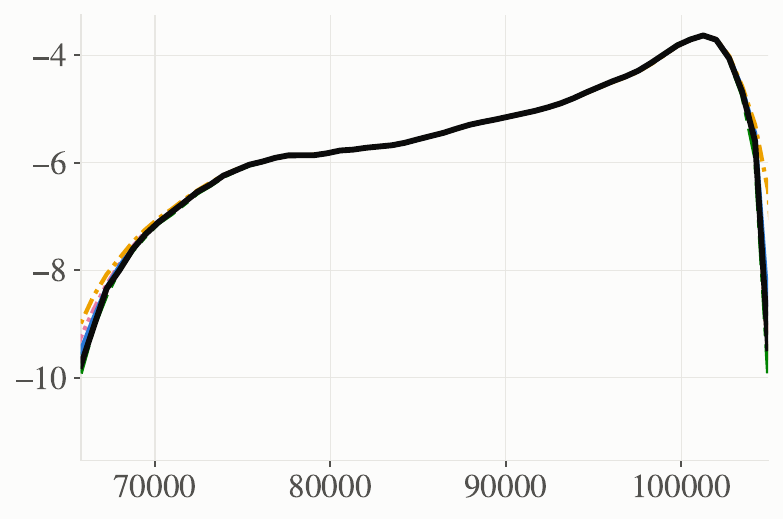}{sp $[\mathrm{Pa}]$}

  \caption{\textbf{Marginal distributions, Central Europe.} $\log_{10}$ of the
  probability density over the test split, so the tails stay visible.}
  \label{fig:logpdf_ce}
\end{figure}

\clearpage

Figures~\ref{fig:psd_iberia} and~\ref{fig:logpdf_iberia} show the same pattern on Iberia. The methods again agree on the smooth variables and separate only on the rough ones (sshf, zust), at the highest wavenumbers and in the tails: bicubic and U-Net lose small-scale energy and contract their marginals, while ours stays closest to the CERRA target, here slightly overshooting its high-frequency energy rather than falling short. As on Central Europe, the large sp error gap between ours and DPS is not visible in either diagnostic, consistent with its being a mean-field bias rather than a spectral or distributional one.

\gridlegend

\begin{figure}[H]
  \centering
  \rowlab{PSD}%
  \gridpanel{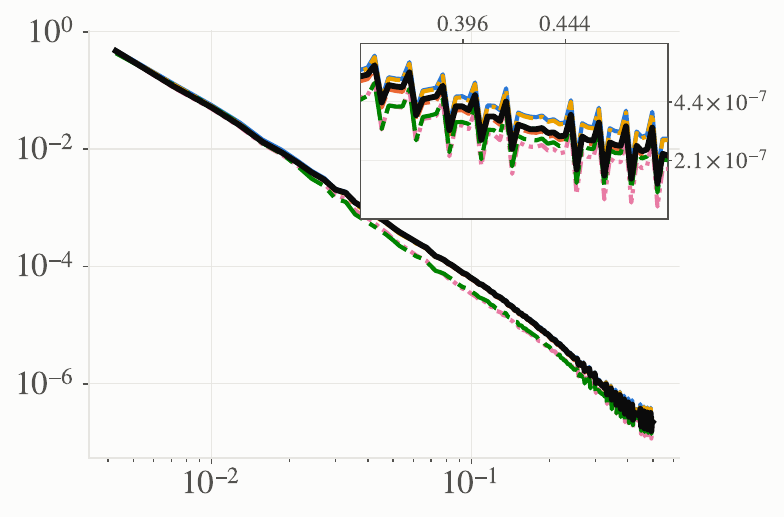}{$u_{10}$ $[\mathrm{m^{2}s^{-2}}]$}\gridgap
  \gridpanel{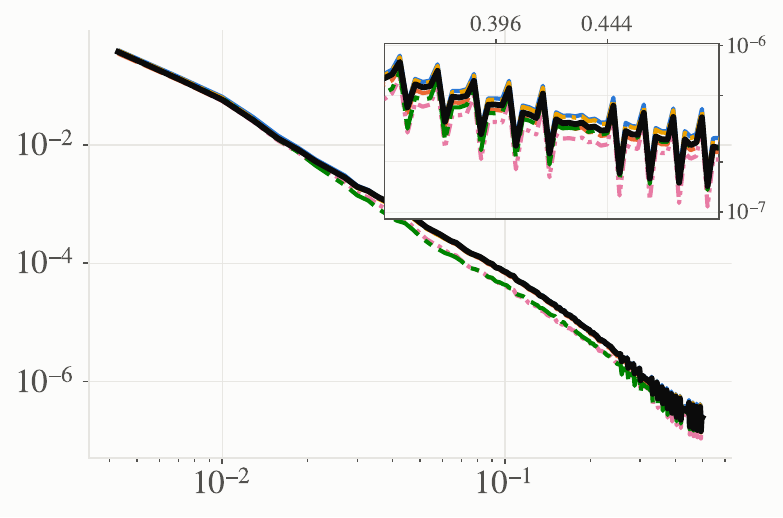}{$v_{10}$ $[\mathrm{m^{2}s^{-2}}]$}\gridgap
  \gridpanel{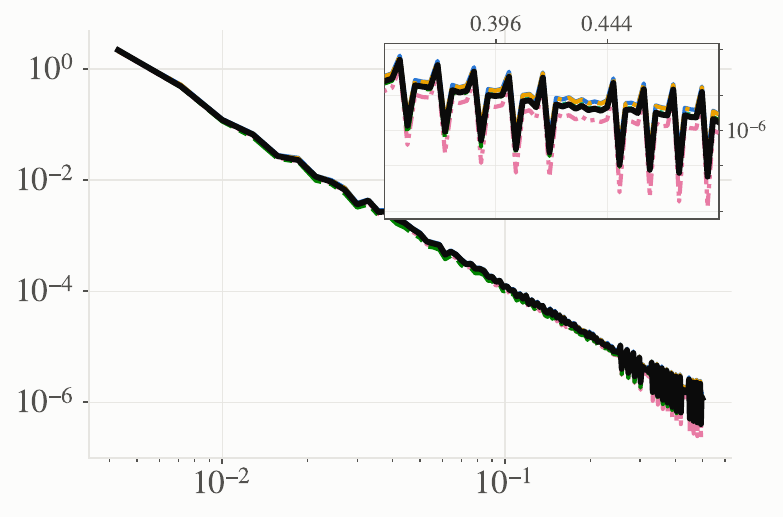}{$t_{2m}$ $[\mathrm{K^{2}}]$}

  \gridrowsep

  \rowlab{PSD}%
  \gridpanel{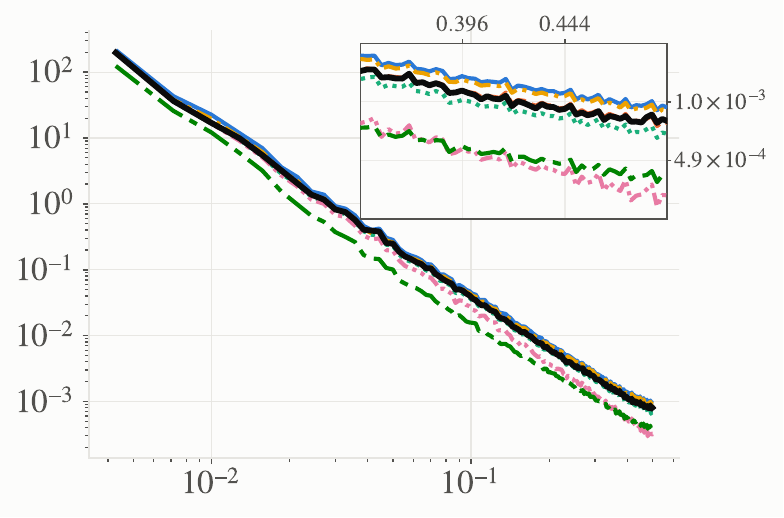}{sshf $[\mathrm{W^{2}m^{-4}}]$}\gridgap
  \gridpanel{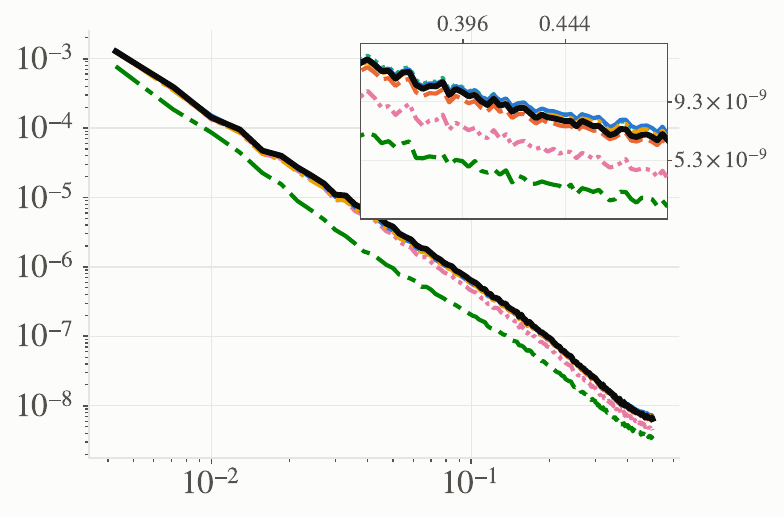}{zust $[\mathrm{m^{2}s^{-2}}]$}\gridgap
  \gridpanel{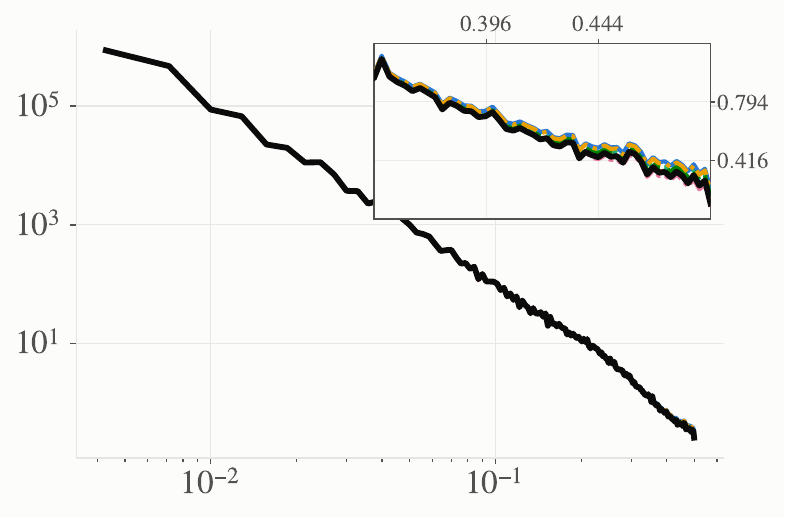}{sp $[\mathrm{Pa^{2}}]$}

  \xlab{wavenumber $k$ $[\mathrm{cycles\;cell^{-1}}]$}

  \caption{\textbf{Power spectral density, Iberia.} Test split, 8-member ensembles. The inset magnifies the highest $0.15$ decades of
  wavenumber, where the curves separate.}
  \label{fig:psd_iberia}
\end{figure}

\begin{figure}[H]
  \centering
  \rowlab{$\log_{10}$ PDF}%
  \gridpanel{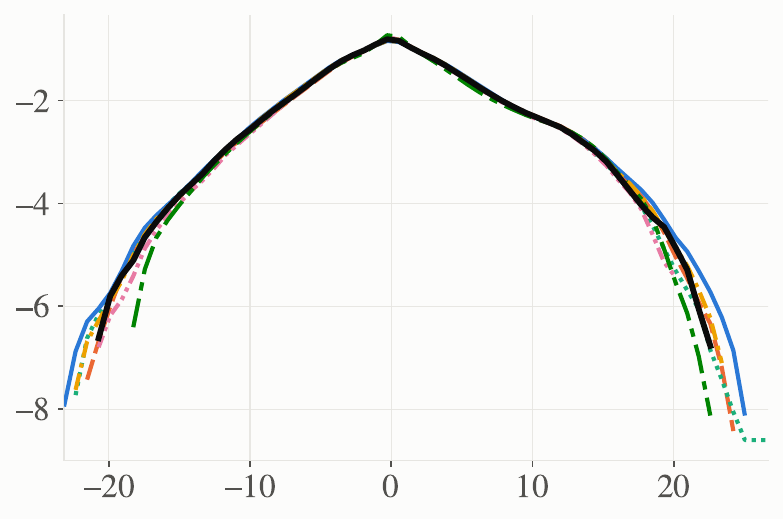}{$u_{10}$ $[\mathrm{m\,s^{-1}}]$}\gridgap
  \gridpanel{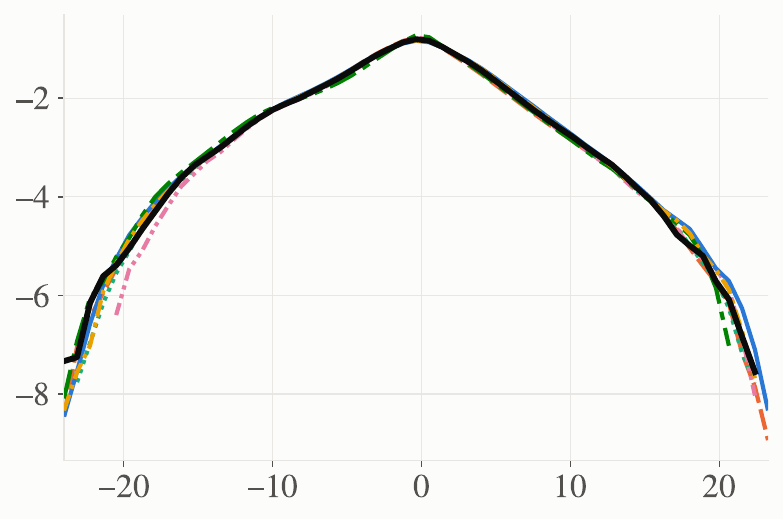}{$v_{10}$ $[\mathrm{m\,s^{-1}}]$}\gridgap
  \gridpanel{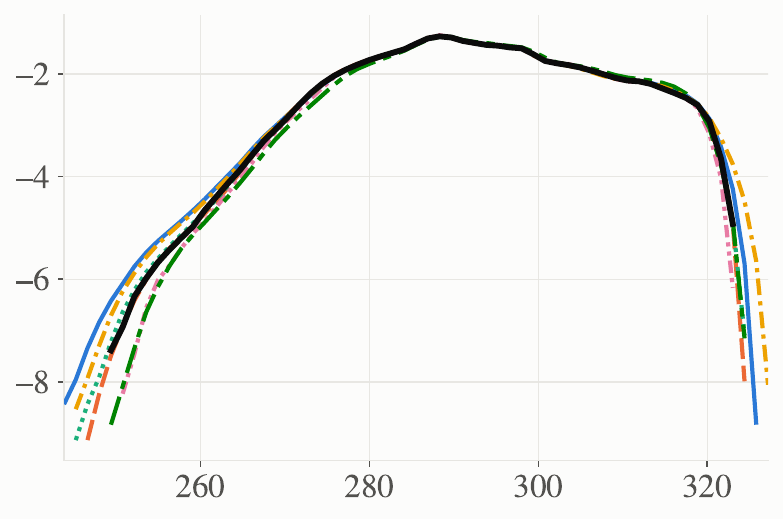}{$t_{2m}$ $[\mathrm{K}]$}

  \gridrowsep

  \rowlab{$\log_{10}$ PDF}%
  \gridpanel{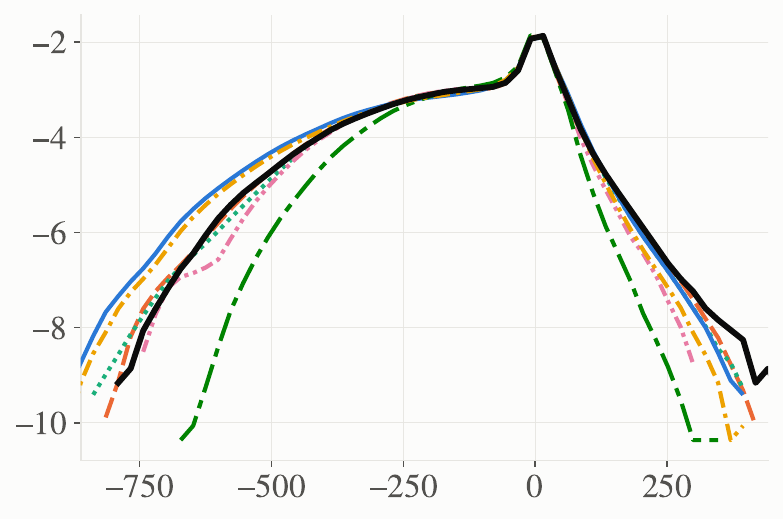}{sshf $[\mathrm{W\,m^{-2}}]$}\gridgap
  \gridpanel{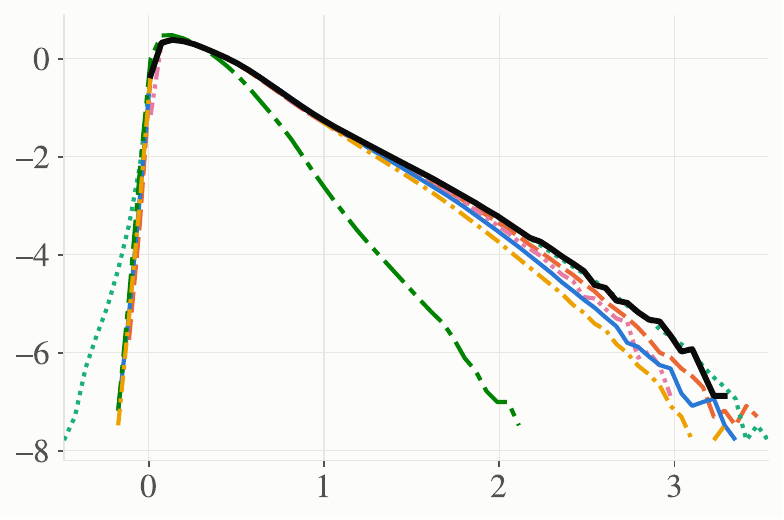}{zust $[\mathrm{m\,s^{-1}}]$}\gridgap
  \gridpanel{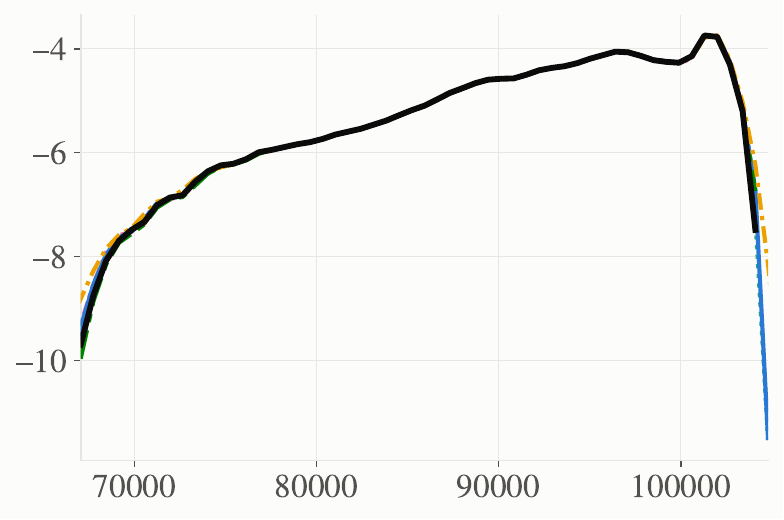}{sp $[\mathrm{Pa}]$}

  \caption{\textbf{Marginal distributions, Iberia.} $\log_{10}$ of the
  probability density over the test split, so the tails stay visible.}
  \label{fig:logpdf_iberia}
\end{figure}

\section{Transfer to a coarser source}
\label{app:transfer}

Adapting to a new coarse source requires retraining only $\mathbf{A}_\phi$, and we test this directly: we halve the resolution of the coarse source to $\sim$$50$\,km, refit $\mathbf{A}_\phi$ on the new pairs, and leave the diffusion prior frozen and unchanged. Everything else in the pipeline is identical.

Table~\ref{tab:transfer} reports the result next to our $\sim$$25$\,km numbers, and Table~\ref{tab:ci_transfer} the corresponding bootstrap intervals. Accuracy is essentially unchanged on both regions: every difference is smaller than the combined intervals of the two estimates. Even with half the input resolution, the method keeps a large margin on sp over both the bicubic-operator and DPS arms at full resolution ($95.68$ here, against $131.2$ and $153.7$ on Central Europe). A conditional model cannot be run on this source at all without changing its input dimensions and retraining in full.

\begin{table}[H]
\centering
\caption{\textbf{Transfer to a $\sim$$50$\,km coarse source.} RMSE and CRPS per variable, both regions. Only $\mathbf{A}_\phi$ is refit; the prior is unchanged. The $\sim$$25$\,km rows repeat Ours from Tables~\ref{tab:baselines_ce} and~\ref{tab:baselines_ib}.}
\label{tab:transfer}
\scriptsize
\setlength{\tabcolsep}{3pt}
\begin{tabular}{ll cc cc cc cc cc cc}
\toprule
\multirow{2}{*}{Region} & \multirow{2}{*}{Coarse source} & \multicolumn{2}{c}{$u_{10}$} & \multicolumn{2}{c}{$v_{10}$} & \multicolumn{2}{c}{$t_{2m}$} & \multicolumn{2}{c}{sshf} & \multicolumn{2}{c}{zust} & \multicolumn{2}{c}{sp} \\
\cmidrule(lr){3-4} \cmidrule(lr){5-6} \cmidrule(lr){7-8} \cmidrule(lr){9-10} \cmidrule(lr){11-12} \cmidrule(lr){13-14}
 & & RMSE & CRPS & RMSE & CRPS & RMSE & CRPS & RMSE & CRPS & RMSE & CRPS & RMSE & CRPS \\
\midrule
\multirow{2}{*}{C.\ Europe}
  & ERA5 $\sim$$25$\,km & 0.986 & 0.553 & 0.982 & 0.547 & 1.021 & 0.571 & 32.70 & 13.33 & 0.090 & 0.047 & 92.05 & 55.96 \\
  & ERA5 $\sim$$50$\,km & 0.994 & 0.557 & 0.989 & 0.551 & 1.055 & 0.590 & 36.79 & 14.80 & 0.091 & 0.047 & 95.68 & 58.85 \\
\midrule
\multirow{2}{*}{Iberia}
  & ERA5 $\sim$$25$\,km & 1.197 & 0.678 & 1.217 & 0.684 & 1.377 & 0.824 & 40.23 & 18.28 & 0.096 & 0.052 & 106.7 & 63.28 \\
  & ERA5 $\sim$$50$\,km & 1.207 & 0.683 & 1.224 & 0.686 & 1.403 & 0.842 & 42.49 & 19.31 & 0.098 & 0.053 & 107.3 & 63.73 \\
\bottomrule
\end{tabular}
\end{table}

\begin{table}[H]
\centering
\caption{\textbf{Transfer to a $\sim$$50$\,km coarse source: bootstrap intervals.} Each entry is the half-width ($\pm$) of the $95\%$ interval for the matching entry in Table~\ref{tab:transfer}. Resampling as in Table~\ref{tab:ci_ce}. The intervals for the $\sim$$25$\,km rows are the Ours rows of Tables~\ref{tab:ci_ce} and~\ref{tab:ci_ib}.}
\label{tab:ci_transfer}
\scriptsize
\setlength{\tabcolsep}{3pt}
\begin{tabular}{ll cc cc cc cc cc cc}
\toprule
\multirow{2}{*}{Region} & \multirow{2}{*}{Coarse source} & \multicolumn{2}{c}{$u_{10}$} & \multicolumn{2}{c}{$v_{10}$} & \multicolumn{2}{c}{$t_{2m}$} & \multicolumn{2}{c}{sshf} & \multicolumn{2}{c}{zust} & \multicolumn{2}{c}{sp} \\
\cmidrule(lr){3-4} \cmidrule(lr){5-6} \cmidrule(lr){7-8} \cmidrule(lr){9-10} \cmidrule(lr){11-12} \cmidrule(lr){13-14}
 & & RMSE & CRPS & RMSE & CRPS & RMSE & CRPS & RMSE & CRPS & RMSE & CRPS & RMSE & CRPS \\
\midrule
\multirow{2}{*}{C.\ Europe}
  & ERA5 $\sim$$25$\,km & 0.022 & 0.011 & 0.021 & 0.010 & 0.032 & 0.019 & 3.65 & 1.09 & 0.0023 & 0.0011 & 3.62 & 2.46 \\
  & ERA5 $\sim$$50$\,km & 0.025 & 0.013 & 0.022 & 0.010 & 0.041 & 0.024 & 4.29 & 1.25 & 0.0027 & 0.0013 & 5.08 & 3.57 \\
\midrule
\multirow{2}{*}{Iberia}
  & ERA5 $\sim$$25$\,km & 0.026 & 0.014 & 0.028 & 0.014 & 0.031 & 0.020 & 2.74 & 1.18 & 0.0023 & 0.0012 & 5.24 & 3.43 \\
  & ERA5 $\sim$$50$\,km & 0.032 & 0.016 & 0.033 & 0.017 & 0.036 & 0.024 & 3.06 & 1.40 & 0.0025 & 0.0013 & 7.18 & 4.70 \\
\bottomrule
\end{tabular}
\end{table}

\newpage
\section{Additional figures}

\begin{figure}[H]
    \centering
    \begin{subfigure}[b]{0.24\linewidth}
        \centering
        \includegraphics[width=\linewidth]{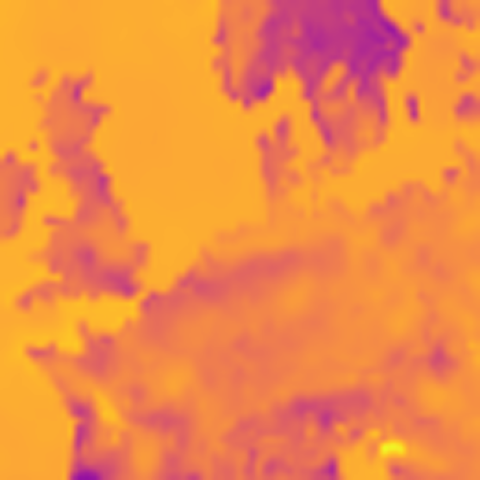}
        \caption{ERA5 Input}
        \label{fig:fixed_A_era5}
    \end{subfigure}%
    \hfill
    \begin{subfigure}[b]{0.24\linewidth}
        \centering
        \includegraphics[width=\linewidth]{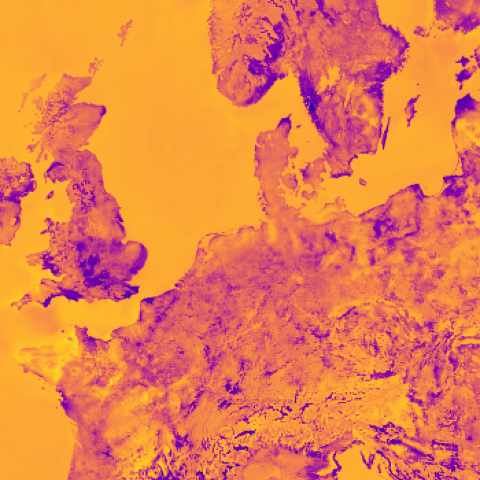}
        \caption{CERRA Target}
        \label{fig:fixed_A_cerra}
    \end{subfigure}%
    \hfill
    \begin{subfigure}[b]{0.24\linewidth}
        \centering
        \includegraphics[width=\linewidth]{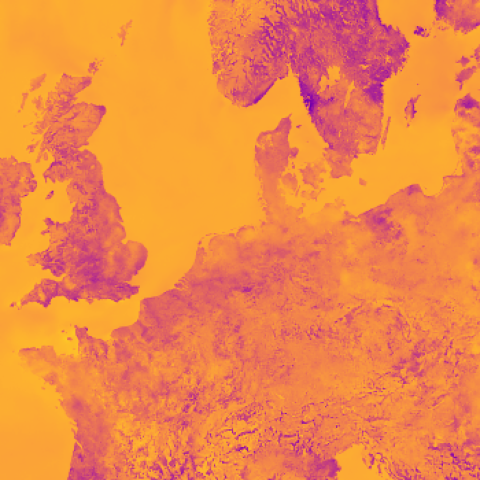}
        \caption{Fixed Operator $\mathbf{A}$}
        \label{fig:fixed_A_pred}
    \end{subfigure}%
    \hfill
    \begin{subfigure}[b]{0.24\linewidth}
        \centering
        \includegraphics[width=\linewidth]{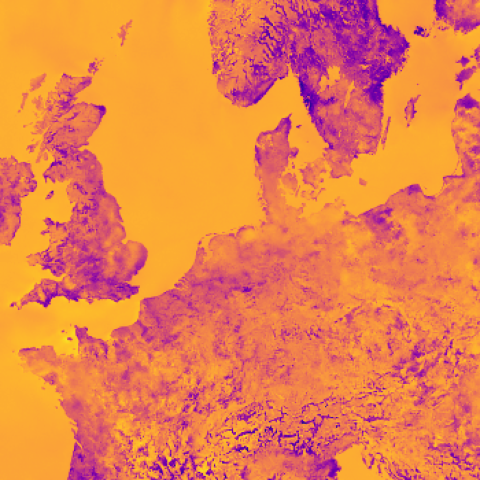}
        \caption{Learned operator (ours)}
        \label{fig:proposed_pred}
    \end{subfigure}
    \caption{\textbf{sshf on a representative sample, Central Europe.} (a) coarse ERA5 input, (b) CERRA target, (c) DPS with a fixed linear operator, (d) our learned operator.}
    \label{fig:fixed_A_failure}
\end{figure}

Figure~\ref{fig:fixed_A_failure}: guided by a fixed linear operator, DPS (c) conflates the reconstruction residual with the ERA5-to-CERRA system bias and leaves visible artefacts, whereas our learned operator (d) recovers a cleaner field close to the CERRA target (b).

\begin{figure}[H]
  \centering
  \includegraphics[width=\linewidth]{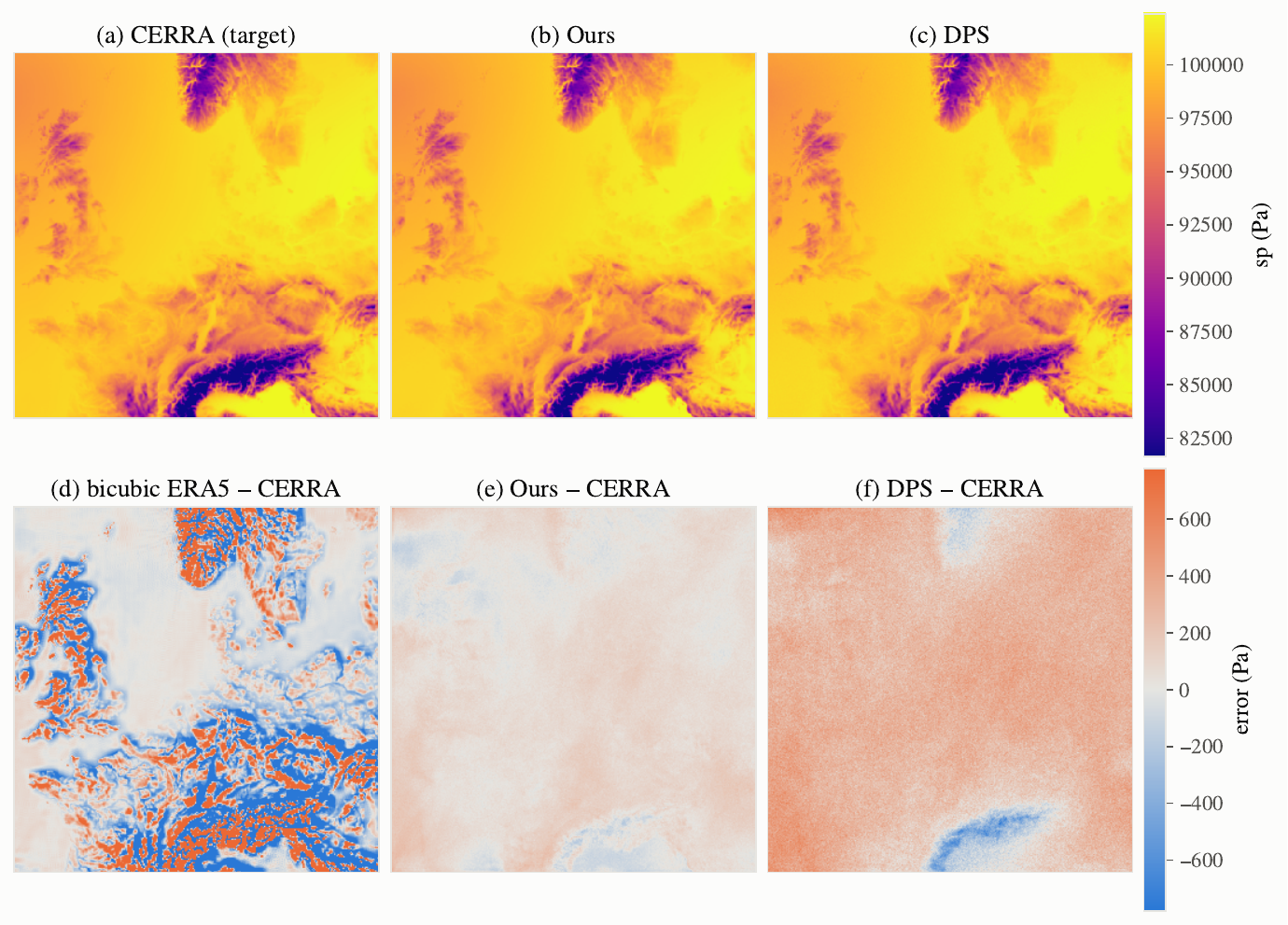}
  \caption{\textbf{Surface pressure on a single test timestep, Central Europe.} Top: CERRA target and the $8$-member ensemble mean from each method. Bottom: error fields on a common scale; (d) bicubic ERA5 residual, (e) ours, (f) DPS.}
  \label{fig:sp_maps_ce}
\end{figure}

Figure~\ref{fig:sp_maps_ce}: ours and DPS share the same prior and learned operator and differ only in gradient scaling. The bicubic residual (d) is an order of magnitude larger than either model error ($1028$\,Pa RMSE, vs.\ $78.6$ ours and $288.2$ DPS) and locked to terrain. The DPS error (f) is a broad one-sign offset ($+250.4$\,Pa domain mean, vs.\ $+54.4$ ours): the field sits at the wrong pressure level rather than the wrong pattern.

\end{document}